\documentclass{article}
\PassOptionsToPackage{numbers, compress}{natbib}
\usepackage[preprint]{neurips_2024}

\usepackage[utf8]{inputenc}
\usepackage[T1]{fontenc}
\usepackage[hidelinks]{hyperref}
\usepackage{url}
\usepackage{booktabs}
\usepackage{amsfonts}
\usepackage{nicefrac}
\usepackage{microtype}
\usepackage{xcolor}
\usepackage{amsmath, bm}
\usepackage{graphicx}
\usepackage{algorithm}
\usepackage{algpseudocode}
\usepackage{tabularx}
\usepackage{multirow}
\usepackage{multicol}
\usepackage{enumitem}
\usepackage{chngcntr}
\usepackage[title]{appendix}
\setlist{leftmargin=4mm}

\DeclareMathOperator{\argmax}{arg\,max}

\title{CONFINE: Conformal Prediction for Interpretable Neural Networks}

\author{
  Linhui Huang$^{1}$ \qquad Sayeri Lala$^{2}$ \qquad Niraj K. Jha$^{2}$ \\
  $^1$Department of Computer Science, Princeton University\\
  $^2$Department of Electrical and Computer Engineering, Princeton University\\
  Princeton, NJ 08540 \\
  \texttt{\{lh9998,slala,jha\}@princeton.edu}
}

\begin{document}
\maketitle

\begin{abstract}
Deep neural networks exhibit remarkable performance, yet their black-box nature limits their utility in fields like healthcare where interpretability is crucial. Existing explainability approaches often sacrifice accuracy and lack quantifiable measures of prediction uncertainty. In this study, we introduce \textbf{Con}formal Prediction \textbf{f}or \textbf{I}nterpretable \textbf{N}eural N\textbf{e}tworks (CONFINE), a versatile framework that generates prediction sets with statistically robust uncertainty estimates instead of point predictions to enhance model transparency and reliability. CONFINE not only provides example-based explanations and confidence estimates for individual predictions but also boosts accuracy by up to 3.6\%. We define a new metric, correct efficiency, to evaluate the fraction of prediction sets that contain precisely the correct label and show that CONFINE achieves correct efficiency of up to 3.3\% higher than the original accuracy, matching or exceeding prior methods. CONFINE's marginal and class-conditional coverages attest to its validity across tasks spanning medical image classification to language understanding. Being adaptable to any pre-trained classifier, CONFINE marks a significant advance towards transparent and trustworthy deep learning applications in critical domains.
\end{abstract}

\section{Introduction}
Deep learning models demonstrate exceptional performance in tasks ranging from medical diagnosis to natural language processing. However, unlike many traditional machine learning models that are easily interpretable, the black-box nature of neural networks makes it a challenge to understand their decision-making process, hindering their adoption in many crucial areas. In fields such as healthcare, where decisions directly affect human lives, interpretability and trustworthiness are essential. 

Many existing explainable Artificial Intelligence (AI) approaches involve modifications to the model architecture or loss functions to obtain explanations in the form of examples or features, which can unfortunately lead to decreased accuracy \citep{Viergever_2022}. Furthermore, many of them are model-specific and none of them provides quantifications to validate the correctness of the explanations.

Explanations indicate why a prediction is made, whereas uncertainty estimates provide an evaluation of the reliability of the prediction. Both aspects are important for interpretability. For example, in a clinical setting, if a model outputs predictions
along with certainty estimates, the clinician can make informed decisions to either trust the most likely diagnosis if the model is certain or order additional diagnostic tests and consider other factors like the patient’s medical history if the model’s confidence is low \citep{aitool}. However, current explainable AI methods focus on elucidating the reasoning behind predictions but fail to provide uncertainty quantification. Simply using softmax scores as confidence estimates is also not robust because adversarial inputs enable crafting of softmax scores that lead to high-confidence incorrect predictions \citep{szegedy2014intriguing}.

\begin{figure}[tbp]
\vskip 0.2in
\begin{center}
\centerline{\includegraphics[width=\linewidth]{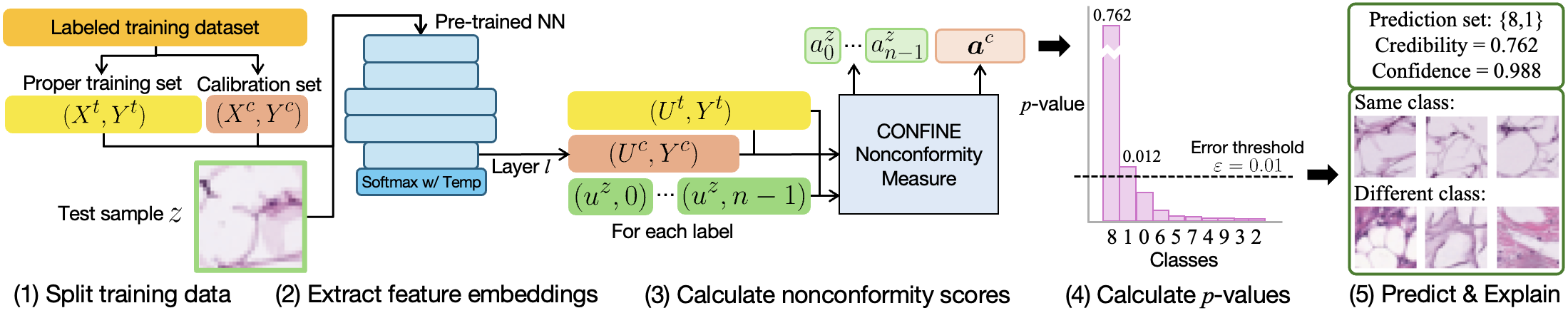}}
\caption{CONFINE applies conformal prediction to enhance the interpretability of any neural network classifier by providing prediction sets with confidence estimates and example-based explanations.}
\label{flowchart}
\vspace{-5px}
\end{center}
\vskip -0.2in
\end{figure}

In this study, we introduce CONFINE, \textbf{Con}formal Prediction \textbf{f}or \textbf{I}nterpretable \textbf{N}eural N\textbf{e}tworks (Fig.~\ref{flowchart}), that leverages conformal prediction \citep{vovk2005} to enhance interpretability by integrating explanations with uncertainty quantification. Our main contributions are as follows:
\begin{itemize}
    \item \textbf{Enhanced Reliability and Transparency}: CONFINE adapts conformal prediction to neural networks to improve model reliability and interpretability. It provides prediction sets with credibility and confidence scores, which help identify ambiguous instances, and example-based explanations that aid in a transparent decision-making process. Compared to prior conformal prediction methods, CONFINE effectively reduces data noisiness through its single-layer feature extraction and top-$k$ nearest neighbors nonconformity measure.
    \item \textbf{Improved Accuracy and Efficiency}: By inducing interpretability, CONFINE can improve the original model's accuracy by up to 3.6\%. We introduce a new metric, called correct efficiency, and demonstrate that CONFINE achieves a correct efficiency up to 3.3\% higher than the original model’s accuracy, matching or outperforming existing methods.
    \item \textbf{Broad Applicability}: CONFINE demonstrates the ability to achieve coverage requirements across various tasks including non-exchangeable patient-based datasets. It can be adapted to any pre-trained machine learning classifier without the need for re-training, thus ensuring broad applicability and ease of integration into existing systems. The code will be made available upon article acceptance.
\end{itemize}

The article is organized as follows.
Section~\ref{sec:background} presents related works on explainable AI, uncertainty estimation, and conformal prediction. Section~\ref{sec:methods} introduces CONFINE's methodology. Section~\ref{sec:experiments} discusses tasks, baselines, and evaluation metrics. Section~\ref{sec:results} presents the experimental results. Section~\ref{sec:limitations} discusses the limitations. Section~\ref{sec:conclusion} summarizes the implications of CONFINE.

\section{Related Works}\label{sec:background}

\subsection{Explainable AI (XAI)}
Existing XAI methods mainly use two forms of explanations: examples and features. Example-based approaches usually offer explanations through optimization of the latent space to highlight analogies or contrasting cases \citep{hoffer2018deep, koh2020understanding, Viergever_2022}. Prototypical approaches \citep{angelov2019explainable, chen2019looks} add prototype layers to obtain image explanations. \citet{Silva_Fernandes_Cardoso_Cardoso_2018} provide similar and dissimilar cases for medical images using nearest neighbor search in the latent space of a deep neural network. Conversely, feature-based approaches focus on producing saliency maps that highlight the important features through techniques like backpropagation or perturbation \citep{jetley2018learn, lundberg2017unified, Viergever_2022}. Early works  \citep{springenberg2015striving, zeiler2013visualizing} visualize feature patterns learned by convolutional layers in a convolutional neural network (CNN). Methods like Class Activation Mapping (CAM) \citep{zhou2015learning}, Grad-CAM \citep{Selvaraju_2019}, Layer-wise Relevance Propagation (LRP) \citep{Bach_Binder_Montavon_Klauschen_Müller_Samek_2015}, and DeepLIFT \citep{shrikumar2019learning} add special layers and look at gradients and activations to produce feature relevance maps. In addition, perturbation-based approaches \citep{Fong_2017, ribeiro2016modelagnostic, zintgraf2017visualizing} perturb the input and look at how the output changes to reveal the specific areas of the input that guide the model's decision. However, most XAI methods suffer from a trade-off between explainability and accuracy because they involve adding or replacing layers in the network, modifying the loss function, or training additional networks. Many XAI methods are model-specific. For example, deconvolution \citep{zeiler2013visualizing}, backpropagation \citep{springenberg2015striving}, CAM \citep{zhou2015learning}, and GradCAM \citep{Selvaraju_2019} are applicable to CNNs only. XAI methods also lack quantifiable evaluation of the reliability of the prediction; hence it is often hard to confirm the correctness of the explanations.

\subsection{Uncertainty Estimation}

In addition to explaining why the model makes a prediction, it is important to provide information about the reliability of individual predictions through uncertainty estimations. There are two sources of uncertainty in machine learning: \textit{aleatoric} and \textit{epistemic}. Previous uncertainty estimation methods mainly leverage Bayesian inference and ensemble learning to provide a probabilistic measure of confidence \citep{Abdar_Pourpanah_Hussain_Rezazadegan_Liu_Ghavamzadeh_Fieguth_Cao_Khosravi_Acharya_et_2021, Zou_Chen_Yuan_Shen_Wang_Fu_2023}. Bayesian Neural Networks \citep{cobb2020scaling, gal2016bayesian, Jospin_2022, Kwon_Won_Kim_Paik_2020, Neal_1996} employ prior distributions on neural network parameters and use the posterior distributions to quantify uncertainty. These posterior distributions are intractable due to nonlinear activations in the network \citep{sensoy2018evidential}; hence, approximation methods, such as Monte Carlo dropout \citep{gal2016bayesian, pmlr-v48-gal16}, are used. Ensemble learning methods aggregate the predictions from an ensemble of models and use their variance as a measure of uncertainty \citep{ashukha2021pitfalls, Lakshminarayanan_Pritzel_Blundell_2017, lee2020ensemble, zhang2020mixnmatch}. These uncertainty estimation methods all require substantial modifications to the model architecture or training procedure, thus adversely impacting applicability to pre-trained models.

\subsection{Conformal Prediction}
Conformal prediction \citep{vovk2005} is a general framework for quantifying uncertainty in predictions. It uses a \textit{nonconformity measure} that indicates the unusualness of a test sample to obtain the likelihood of it being in a certain class. It can be applied to any prediction algorithm through the re-design of the nonconformity measure. 

Conformal prediction has been adapted to various traditional machine learning models, like linear regression, support vector machines, and nearest neighbors \citep{shafer2007tutorial}, as well as neural networks \citep{confnn, papernot2018deep}. The 1-Nearest Neighbor conformal predictor \citep{vovk2005} uses distances to the nearest same-class and different-class training samples in the nonconformity measure,
\begin{equation} \label{1nn}
    A(B,x,y) := \frac{\min_{y_i=y} d(x_i,x)}{\min_{y_i\neq y} d(x_i,x)}, (x_i,y_i)\in B,
\end{equation}
where $d(x_i,x)$ refers to the cosine distance between the input vectors of sample $x_i\in B$ and the test sample $x$. However, using nearest neighbors can be inefficient, particularly when dealing with high-dimensional data and large training sets, and focusing solely on the top nearest neighbor introduces noise. Neural Network Inductive Conformal Prediction (ICP) \citep{confnn} uses two nonconformity measures based on softmax scores. A sample's nonconformity score is given by\vspace{5px}
\noindent\begin{minipage}{.45\linewidth}
\begin{equation}\label{noncon1}
    \alpha := \max_{j=0,..,n-1:j\neq u} o_j - o_u,
\end{equation}
\end{minipage}
\begin{minipage}{.55\linewidth}
\begin{equation}\label{noncon2}
    \alpha := \frac{\max_{j=0,..,n-1:j\neq u} o_j}{o_u+\gamma},\gamma\geq 0,
\end{equation}
\end{minipage}\vspace{5px}
where $o_j$ refers to the softmax score of class $j$, $u$ is the class with the highest softmax score, and $\gamma$ is a hyperparameter that enhances stability. These softmax-based nonconformity measures maintain the original network’s accuracy without incurring extra time or storage. However, they lack robustness -- adversarial samples erroneously display high softmax scores on incorrect labels \citep{szegedy2014intriguing}. Deep $k$-Nearest Neighbors (D$k$NN) \citep{papernot2018deep} employs $k$-nearest neighbors on outputs from every layer of the network, using the fraction of same-class neighbors as the nonconformity measure. However, this method leads to accuracy drops and is too computationally expensive for deep networks. Our method, CONFINE, addresses these challenges by selecting a single layer for feature extraction to reduce the computational burden and considering top-$k$ nearest neighbors to avoid noisiness in the training data.

\section{The CONFINE Framework}\label{sec:methods}
We begin with a review of conformal prediction \citep{vovk2005}, then explain how CONFINE adapts conformal prediction to neural networks to provide both uncertainty scores and explanations.

\subsection{Review of Conformal Prediction}
Given an exchangeable dataset $((x_1,y_1),...,(x_l,y_l))$ and a test sample $x_{l+1}$, conformal prediction calculates the likelihood of $x_{l+1}$ being in each potential class $Y_j$ ($Y_j\in\mathcal{C}$, $\mathcal{C}$ is the set of all classes) by measuring the typicalness of the extended sequence $((x_1,y_1),...,(x_l,y_l),(x_{l+1},Y_j))$ \citep{Papadopoulos_2008}. This likelihood estimate, $p(Y_j)$, is called the $p$-value of $Y_j$ because it adheres to the statistical definition of $p$-values,
\begin{equation} \label{pvalue}
    \Pr[p(Y_j) \leq \delta] \leq \delta, \forall\delta\in[0,1].
\end{equation}
A \textit{nonconformity measure} $A(B,z)$ is a family of functions that takes in a set of labeled examples $B$ and an example with a label $z=(x,y)$ and assigns a numerical score. By using a nonconformity measure to determine how unusual each sample $(x_i,y_i)$ looks relative to all others in the sequence,
\begin{equation} \label{noncon}
    \alpha_i = A([z_1,...,z_{i-1},z_{i+1},...,z_{l+1}],z_i), \text{where }z_i=(x_i,y_i),
\end{equation}
the $p$-value of $Y_j$ is then calculated as the fraction of other samples with a nonconformity score higher than or equal to that of the test sample,
\begin{equation} \label{pyj}
\begin{split}
    p(Y_j) = \frac{\#\{i=1,...,l+1: \alpha_i \geq \alpha_{l+1}\}}{l+1}.
\end{split}
\end{equation}
A proof of this $p$-value satisfying Eq.~(\ref{pvalue}) can be found in \citep{Nouretdinov_Vovk_Vyugin_Gammerman_2001}. Repeating this calculation for all possible classes gives the $p$-values of the test sample $x_{l+1}$ belonging to all possible labels, $p(Y_j), Y_j\in\mathcal{C}$. Using these $p$-values, conformal prediction generates a prediction set $\Gamma^\varepsilon$ containing all classes with $p$-values higher than a user-specified significance level $\varepsilon$,
\begin{equation}
    \Gamma^\varepsilon := \{Y_j: p(Y_j) > \varepsilon\},\forall Y_j\in\mathcal{C}.
\end{equation}
Credibility and confidence of the prediction are also calculated using the $p$-values,
\begin{align} \label{credibility}
    \text{Credibility} = \max_{Y_j\in\mathcal{C}} p(Y_j), \hspace{4px}
    \text{Confidence} = 1-\max_{Y_j\in\mathcal{C}, Y_j\neq \argmax_{Y_i}p(Y_i)} p(Y_j).
\end{align}
Credibility reflects how suitable the training data are for classifying the new sample. A low credibility indicates insufficient diversity of the training set to cover the region necessary for correctly classifying the test sample. Confidence provides a measure of certainty against the next most likely class.

The prediction set is guaranteed to cover the true label $1-\varepsilon$ of the time, that is,
\begin{equation} \label{marginal-coverage}
    \Pr[y_i\in \Gamma^\varepsilon_i] \geq 1-\varepsilon,
\end{equation}
where $y_i$ is the ground truth label of a test sample $i$, and $\Gamma^\varepsilon_i$ is the prediction set from conformal prediction for sample $i$ with a significance level $\varepsilon$. This is called the \textit{marginal coverage}. Conformal prediction's coverage is guaranteed under the \textit{exchangeability assumption}, which states that the data distribution is invariant under permutations. It is a relaxation of the i.i.d. assumption.

The more desirable \textit{class-conditional coverage} is defined as 
\begin{equation} \label{cc-coverage}
    \Pr[y_i\in \Gamma^\varepsilon_i | y_i=Y_j] \geq 1-\varepsilon, \forall Y_j\in \mathcal{C},
\end{equation}
where $\mathcal{C}$ is the set of all possible classes. This means that every class $Y_j$ is included in the prediction set with probability $1-\varepsilon$ when $Y_j$ is the true label. Class-conditional coverage can be achieved by splitting the training data by class and obtaining the $p$-value of a class only using the training samples from that class \citep{Vovk_2012}. Albeit important, few conformal prediction methods take class-conditional coverage into account.

\begin{algorithm}[tbp]
\small 
\begin{algorithmic}
    \State {\bfseries Inputs:} labeled training dataset $(X,Y)$, pre-trained neural network $f$, significance level $\varepsilon$, test input $z$
    \State {\bfseries Hyperparameters:} layer $l$, number of nearest neighbors $k$, temperature $T$ (if $l=$ softmax)
\end{algorithmic}
\begin{algorithmic}[1]
    \State Split dataset $(X,Y)$ into proper training set $(X^t,Y^t)$ and calibration set $(X^c,Y^c)$
    \State Extract feature embeddings from layer $l$ of the neural network
    \vspace{-3px}
    \begin{align*}
        U^t := \{f_l(x_i) | x_i\in X^t\}, \hspace{3px} U^c := \{f_l(x_i) | x_i\in X^c\}, \hspace{3px} u := f_l(z)
    \end{align*}
    \vspace{-12px}
    \State Calculate nonconformity scores for calibration set using CONFINE nonconformity measure (Eq.~(\ref{confine_noncon_measure}))
    \vspace{-4px}
    \[\bm{\alpha}^c := \{A_k((U^t,Y^t), u_i, y_i) | (u_i,y_i)\in (U^c,Y^c)\}\]
    \vspace{-14px}
    \For{each class $y_j\in 0,..,n-1$}
        \State Calculate nonconformity score and $p$-value of test sample $z$: $\alpha(z,y_j) = A_k((U^t,Y^t), u, y_j)$
        \begin{itemize}
        \item CONFINE: using all calibration samples 
        \vspace{-6px}
        \[p(y_j) = \frac{|\{\alpha_i^c\in \bm{\alpha}^c: \alpha_i^c \geq \alpha(z,y_j)\}|+1}{|\bm{\alpha}^c|+1}\]
        \vspace{-9px}
        \item CONFINE-classwise: using calibration samples from class $y_j$ only
        \vspace{-4px}
        \[p(y_j) = \frac{|\{\alpha_i^c\in \bm{\alpha}^c: Y_i^c = y_j, \alpha_i^c \geq \alpha(z,y_j)\}|+1}{|\bm{\alpha}^c|+1}\]
        \vspace{-9px}
        \end{itemize}
        \State Include $y_j$ in prediction set $\Gamma^\varepsilon$ if and only if $p(y_j) > \varepsilon$
    \EndFor
    \State prediction $\gets \argmax_{y_j\in 0,..,n-1} p(y_j)$
    \State credibility $\gets \max_{y_j\in 0,..,n-1} p(y_j)$. confidence $\gets 1-\max_{y_j\in 0,..,n-1, y_j\neq \text{prediction}} p(y_j)$
    \State {\bfseries return} prediction, confidence, credibility, prediction set $\Gamma^\varepsilon$
\end{algorithmic} 
\caption{The CONFINE Algorithm}
\label{alg:confine}
\end{algorithm}

\subsection{The CONFINE Algorithm}

CONFINE extends conformal prediction to neural networks to provide explanations in addition to the prediction set, credibility, and confidence obtained from the original conformal prediction framework. We first define CONFINE's nonconformity measure using the cosine distances of the top-$k$ nearest neighbors, where $k$ is a hyperparameter, \vspace{-6px}
\begin{align} \label{confine_noncon_measure}
\begin{split}
A_k(B,x,y) &:= \frac{\min\{\frac{1}{k}\sum_{i=1}^k \mathrm{CosDist}(x, x_i), y_i=y\}}{\min\{\frac{1}{k}\sum_{i=1}^k \mathrm{CosDist}(x, x_i), y_i\neq y\}} = \frac{\begin{tabular}{@{}c@{}}\small
      average cosine distance of top $k$ nearest \\\small
      neighbors in $B$ with the same label
    \end{tabular}}{\begin{tabular}{@{}c@{}}\small
      average cosine distance of top $k$ nearest \\\small
      neighbors in $B$ with a different label
    \end{tabular}}\\
\text{where }&(x_i, y_i)\in B, \mathrm{CosDist}(a, b) := 1 - \frac{a\cdot b}{\|a\| \|b\|}.
\end{split}
\end{align}\vspace{-12px}

The full CONFINE algorithm is described in Fig.~\ref{flowchart} and Algorithm~\ref{alg:confine}. 

\textbf{Data preparation.}\hspace{8px} A labeled training dataset is split into a proper training set and a calibration set. The calibration set,  following the ICP approach \citep{Vovk_2012}, ensures that nonconformity scores need not be recalculated every time the model processes a new test sample. Given a pre-trained machine learning classifier, CONFINE appends a softmax layer with temperature after the last fully-connected layer of the network: $ \mathrm{softmax}_i = \frac{\exp({z_i/T})}{\sum_{j}^{C} \exp(z_j/T)}$, where $C$ is the number of total classes, $z_j$ are the output logits from the network, and temperature $T$ is a hyperparameter.

\textbf{Feature extraction.}\hspace{8px} Feature embeddings for the proper training set, the calibration set, and the test sample are extracted using the outputs from the $l$-th layer of the pre-trained neural network. $l$ is a hyperparameter selected using grid search. The layer right before the fully-connected classification head usually works best for capturing the essential features. 

\textbf{Calculation of nonconformity scores.}\hspace{8px} The nonconformity scores are calculated using CONFINE's top-$k$ nearest neighbors nonconformity measure (Eq.~(\ref{confine_noncon_measure})) both between the calibration set and the proper training set, and between the test sample and the proper training set for each possible class. Proper training samples incorrectly classified by the network are removed to improve accuracy.

\textbf{Calculation of $p$-values.}\hspace{8px} Each label's $p$-value is calculated as the fraction of calibration nonconformity scores higher than or equal to the test sample. To achieve class-conditional coverage (Eq.~(\ref{cc-coverage})), the $p$-value of a label is calculated only using the calibration samples from that class. We report the performance of CONFINE with and without the application of this class-wise split.

\textbf{Prediction and explanation.}\hspace{8px} Lastly, labels with $p$-values higher than the user-specified error threshold $\varepsilon$ are included in the prediction set. Credibility and confidence are calculated using Eq.~(\ref{credibility}). The same-class and different-class top-$k$ nearest neighbors that lead to this prediction are also visualized along with their distances to the test sample. The CONFINE framework thus equips users with detailed insights into the decision-making process, leading to a transparent, interpretable, and reliable machine learning model.

\section{Experimental Setup}\label{sec:experiments}
\subsection{Tasks and Baselines}
We evaluate CONFINE on three distinct tasks: medical sensor data-based disease diagnosis (CovidDeep) \citep{hassantabar2020coviddeep}, image classification (CIFAR-10 \citep{Krizhevsky09learningmultiple} and PathMNIST and OrganAMNIST from MedMNIST \citep{Yang_2023}), and natural language understanding (GLUE SST-2 and MNLI) \citep{wang2019glue}. Dataset and training details are provided in Appendix~\ref{app:tasks}.

We compare CONFINE against four prior methods: 1-Nearest Neighbor \citep{vovk2005} (Eq.~(\ref{1nn})), two nonconformity measures from Neural Networks ICP \citep{confnn} (Eqs.~(\ref{noncon1}), (\ref{noncon2})), and D$k$NN \citep{papernot2018deep}.

\subsection{Evaluation Metrics}
\textbf{Accuracy.}\hspace{7px} Top-1 prediction accuracy is defined as the accuracy of the label with the highest $p$-value.

\textbf{Correct Efficiency.}\hspace{8px} Conformal prediction is efficient only if the prediction region is small and accurate. Hence, given a test set, we define a \textit{Correct Efficiency} metric that measures the fraction of prediction sets that contain exactly the correct label, \vspace{-5px}
\begin{equation}\label{correct_efficiency}
    \mathrm{CorrectEfficiency}(\{\Gamma^\varepsilon\}) = \frac{\big|\big\{\Gamma^\varepsilon_i | \Gamma^\varepsilon_i = \{y_i\}\big\}\big|}{|\{\Gamma^\varepsilon\}|},
\end{equation}
where $\{\Gamma^\varepsilon\}$ represents prediction sets of all test samples, $\Gamma^\varepsilon_i$ is the prediction set of test sample $i$, and $y_i$ is the ground truth label of sample $i$. Since it assesses both accuracy and precision, correct efficiency is crucial for evaluating how effectively a conformal predictor identifies the correct label without overgeneralization. We report the top correct efficiency achieved after a grid search of $\varepsilon$.

\textbf{Coverage.}\hspace{8px} We also evaluate coverage, the fraction of prediction sets that contain the correct label, to assess the validity of CONFINE, \vspace{-5px}
\begin{equation}\label{coverage_eqn}
    \mathrm{Coverage}(\{\Gamma^\varepsilon\}) = \frac{|\{\Gamma^\varepsilon_i | y_i \in \Gamma^\varepsilon_i \}|}{|\{\Gamma^\varepsilon\}|}.
\end{equation}

\begin{figure}[tbp]
    \centering
    \includegraphics[width=\linewidth]{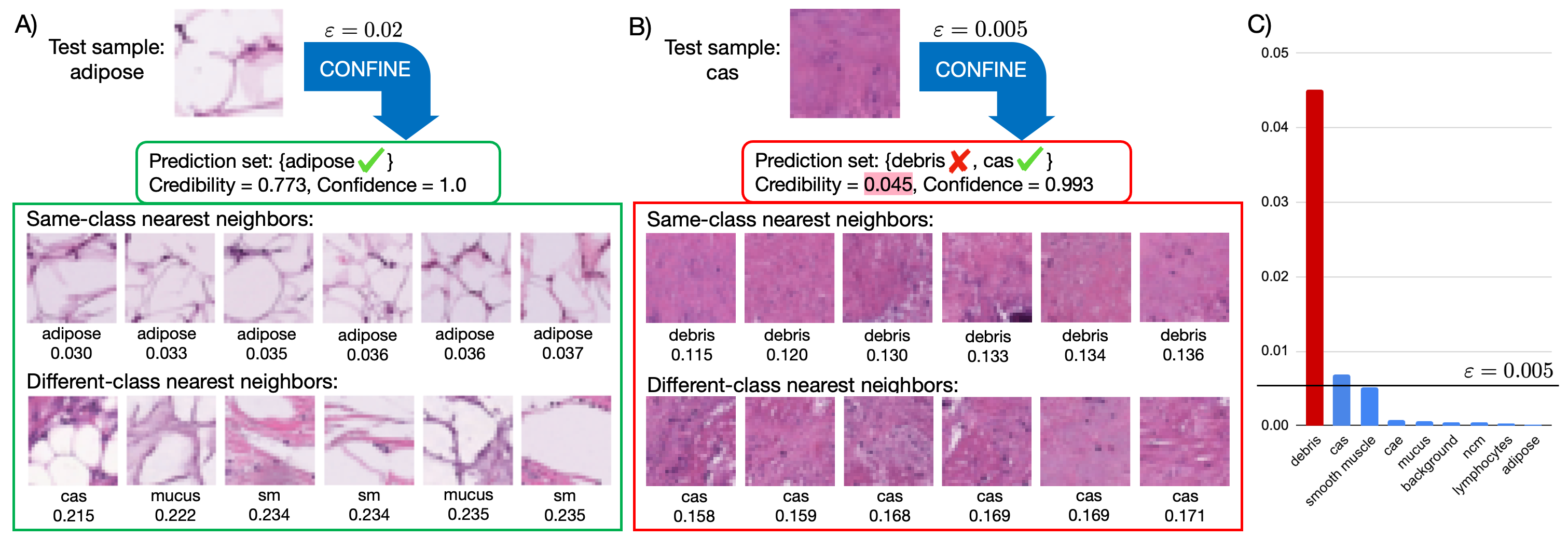}
    \vspace*{-15px}
    \caption{Two test samples from PathMNIST: (A) correctly classified by CONFINE and (B) incorrectly classified by CONFINE. (C) $p$-values for the test sample in (B). Classes with $p$-values above the significance level $\varepsilon$ are included in the prediction set. Hyperparameters used: $k=20, l=50$. cas: cancer-associated stroma, sm: smooth muscle, cae: colorectal adenocarcinoma epithelium, ncm: normal colon mucosa.}
    \label{fig:example}
    \vspace*{-8px}
\end{figure}

\begin{table}[htbp]
\caption{CONFINE achieves the same or higher accuracies and top correct efficiencies than all prior conformal prediction methods. CONFINE: without calibration set class-wise split. -A: hyperparameter search for highest test accuracy; -C: hyperparameter search for highest top correct efficiency. Hyperparameter details are given in Appendix~\ref{app:hparams}. D$k$NN encountered out-of-memory error for all datasets except CovidDeep; hence, its results are not included.}
\vspace*{-24px}
\label{tab:accuracy}
\vskip 0.15in
\begin{center}
\footnotesize
\begin{tabular}{lcccccc}
\toprule
\multirow{2}{*}{Method} & \multicolumn{2}{c}{CIFAR-10} & \multicolumn{2}{c}{PathMNIST} & \multicolumn{2}{c}{OrganAMNIST}\\
\cmidrule(l{2px}r{2px}){2-3} \cmidrule(l{2px}r{2px}){4-5} \cmidrule(l{2px}r{2px}){6-7}
 & Test Acc & Top Corr Eff & Test Acc & Top Corr Eff & Test Acc & Top Corr Eff\\
\midrule
Original NN & 93.92 & -- & 90.65 & -- & 91.78 & --\\
1-Nearest Neighbor (\ref{1nn})  & 40.54 ($\downarrow$) & 40.06 & 35.33 ($\downarrow$) & 34.92 & 71.73 ($\downarrow$) & 70.54 \\
Noncon. Measure 1 (\ref{noncon1}) & 93.92 (--) & \textbf{93.92} & 90.65 (--) & 90.42  & 91.78 (--) & 89.82 \\
Noncon. Measure 2 (\ref{noncon2}) & 93.92 (--) & \textbf{93.92} & 90.65 (--) & 90.46 & 91.78 (--) & 89.88 \\
\textbf{CONFINE-A (ours)} & \textbf{94.37} ($\uparrow$) & 93.72 & \textbf{94.22} ($\uparrow$) & 93.44 & \textbf{94.78} ($\uparrow$) & 92.94 \\
\textbf{CONFINE-C (ours)} & 93.92 (--) & \textbf{93.92} & 94.09 ($\uparrow$) & \textbf{93.91} & 94.45 ($\uparrow$) & \textbf{94.04} \\
\midrule
\multirow{2}{*}{Method} & \multicolumn{2}{c}{CovidDeep} & \multicolumn{2}{c}{SST-2} & \multicolumn{2}{c}{MNLI}\\
\cmidrule(l{2px}r{2px}){2-3} \cmidrule(l{2px}r{2px}){4-5} \cmidrule(l{2px}r{2px}){6-7}
 & Test Acc & Top Corr Eff & Test Acc & Top Corr Eff & Test Acc & Top Corr Eff\\
\midrule
Original NN & \textbf{98.07} & -- & 93.23 & -- & 86.56 & -- \\
1-Nearest Neighbor (\ref{1nn}) & 66.25 ($\downarrow$) & 66.22 & 51.49 ($\downarrow$) & 51.49 & 33.74 ($\downarrow$) & 32.98 \\
D$k$NN-A \citep{papernot2018deep} & 96.41 ($\downarrow$) & 91.00 & --& --& --& --\\
D$k$NN-C \citep{papernot2018deep} & 96.01 ($\downarrow$) & 95.98 & --& --& --& --\\
Noncon. Measure 1 (\ref{noncon1}) & \textbf{98.07} (--) & \textbf{98.07} & 93.23 (--) & 92.32 & 86.56 (--) & 85.68 \\
Noncon. Measure 2 (\ref{noncon2}) & \textbf{98.07} (--) & \textbf{98.07} & 93.23 (--) & 92.55 & 86.56 (--) & 85.92 \\
\textbf{CONFINE-A (ours)} & \textbf{98.07} (--) & \textbf{98.07} & \textbf{93.69} ($\uparrow$) & 92.43 & \textbf{86.67} ($\uparrow$) & 85.71 \\
\textbf{CONFINE-C (ours)} & \textbf{98.07} (--) & \textbf{98.07} & 93.57 ($\uparrow$) & \textbf{92.66} & 86.60 ($\uparrow$) & \textbf{85.79}\\
\bottomrule
\vspace*{-13px}
\end{tabular}
\end{center}
\vskip -0.1in
\end{table}

\vspace{-10px}
\section{Experimental Results}\label{sec:results}

\subsection{CONFINE Provides Interpretability}
CONFINE significantly enhances the interpretability of neural network predictions. Fig.~\ref{fig:example}A illustrates an example when CONFINE successfully predicts the correct label with a high credibility of 0.773 and a high confidence of 1.0. For the correctly predicted label ``adipose,'' the same-class top nearest neighbors are notably closer to the test sample with distances around 0.03, while the different-class neighbors all have distances over 0.2 to the test sample. The nonconformity score calculated using distances to these neighbors leads to a very high $p$-value for ``adipose'' and very low $p$-values for all other classes, explaining why the model is confident and outputs the correct prediction.

Conversely, Fig.~\ref{fig:example}B shows a harder example that CONFINE fails to predict correctly and suggests to the user that it is uncertain. Given the test sample from the class ``cancer-associated stroma,'' CONFINE outputs a prediction set containing ``debris'' and ``cancer-associated stroma'' with a low credibility of 0.045, indicating that the training data possibly lack diversity to cover the latent space well enough to classify this sample. Indeed, both same-class and different-class nearest neighbors look very similar to the test sample. Their distances to the test sample are close in magnitude, at around 0.12 to 0.17, leading to a high nonconformity score and hence relatively low $p$-values for all the labels, as confirmed by the $p$-value distribution in Fig.~\ref{fig:example}C. The true label ``cancer-associated stroma'' is included in the prediction set if $\varepsilon$ is set low enough, e.g., 0.005. Given the prediction set of size larger than 1 and the low credibility, the user would be able to identify that the model is incapable of correctly classifying this particular sample and, thus, should further analyze it.

\subsection{CONFINE Boosts Accuracy}
CONFINE demonstrates a remarkable ability to improve the prediction accuracy of neural networks, matching or outperforming both the original networks and previous conformal prediction methods. Table~\ref{tab:accuracy} shows this enhancement across all six datasets. Taking PathMNIST as an example, CONFINE's integration of interpretability and uncertainty estimation actually increases accuracy by 3.57\% over the original neural network. Note that the two neural network nonconformity measures (Eqs.~(\ref{noncon1}) and (\ref{noncon2})), which rely solely on softmax scores, fail to improve the original neural network's accuracy. Compared with existing XAI methods that either sacrifice or maintain accuracy, CONFINE can refine the predictive capabilities of neural networks through explainability and uncertainty quantification. Since conformal prediction involves no re-training, the result from each experiment is deterministic, and hence no error bars are necessary for the results reported in this paper.

\subsection{CONFINE Boosts Correct Efficiency}
CONFINE excels not only at enhancing prediction accuracy but also at improving correct efficiency, the fraction of prediction sets that precisely contain the correct label. Again, as shown in Table~\ref{tab:accuracy}, CONFINE consistently achieves the same or higher top correct efficiency than all prior methods across all tested datasets. For example, on PathMNIST, CONFINE achieves a correct efficiency that is 3.26\% higher than the original network's accuracy, further demonstrating its effectiveness in not only predicting the correct labels but also doing so with a high degree of confidence and specificity.

\subsection{Coverage Suggests Validity of CONFINE}\label{sec:validity}
CONFINE demonstrates strong performance on coverage metrics, underscoring its validity. Coverage measures the fraction of prediction sets that include the true label. A conformal predictor is valid only when coverage is always higher than $1-\varepsilon$, indicating that it abides by the allowed error rate $\varepsilon$.

\begin{figure}[htbp]
    \centering
    \includegraphics[width=\linewidth]{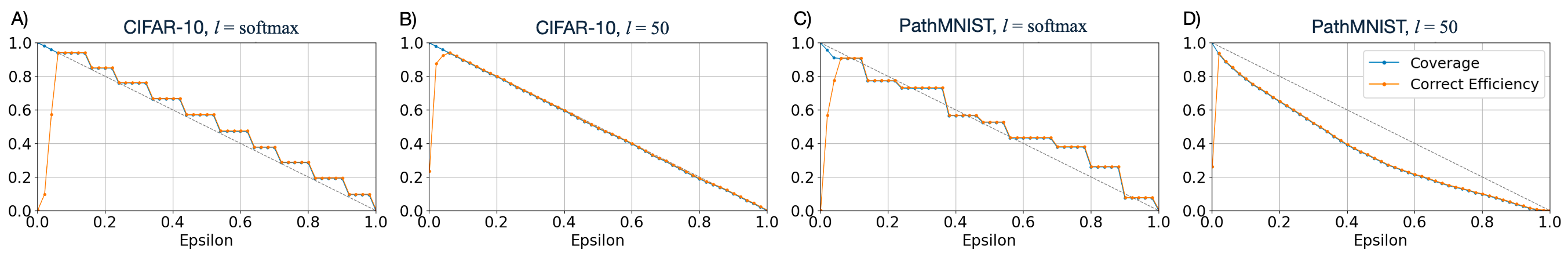}
    \vspace*{-15px}
    \caption{CONFINE's coverage and correct efficiency curves when changing allowed error rate $\varepsilon$. Coverage being above the diagonal line means that the coverage follows the allowed error rate and conformal prediction is valid. Hyperparameters used: (A) $T = 0.01, k = 5$, (B) $k = 5$, (C) $T = 0.01, k = 5$, (D) $k = 20$.}
    \label{fig:coverage}
    \vspace*{-10px}
\end{figure}

\begin{figure*}[htbp]
    \centering
    \includegraphics[width=0.8\linewidth]{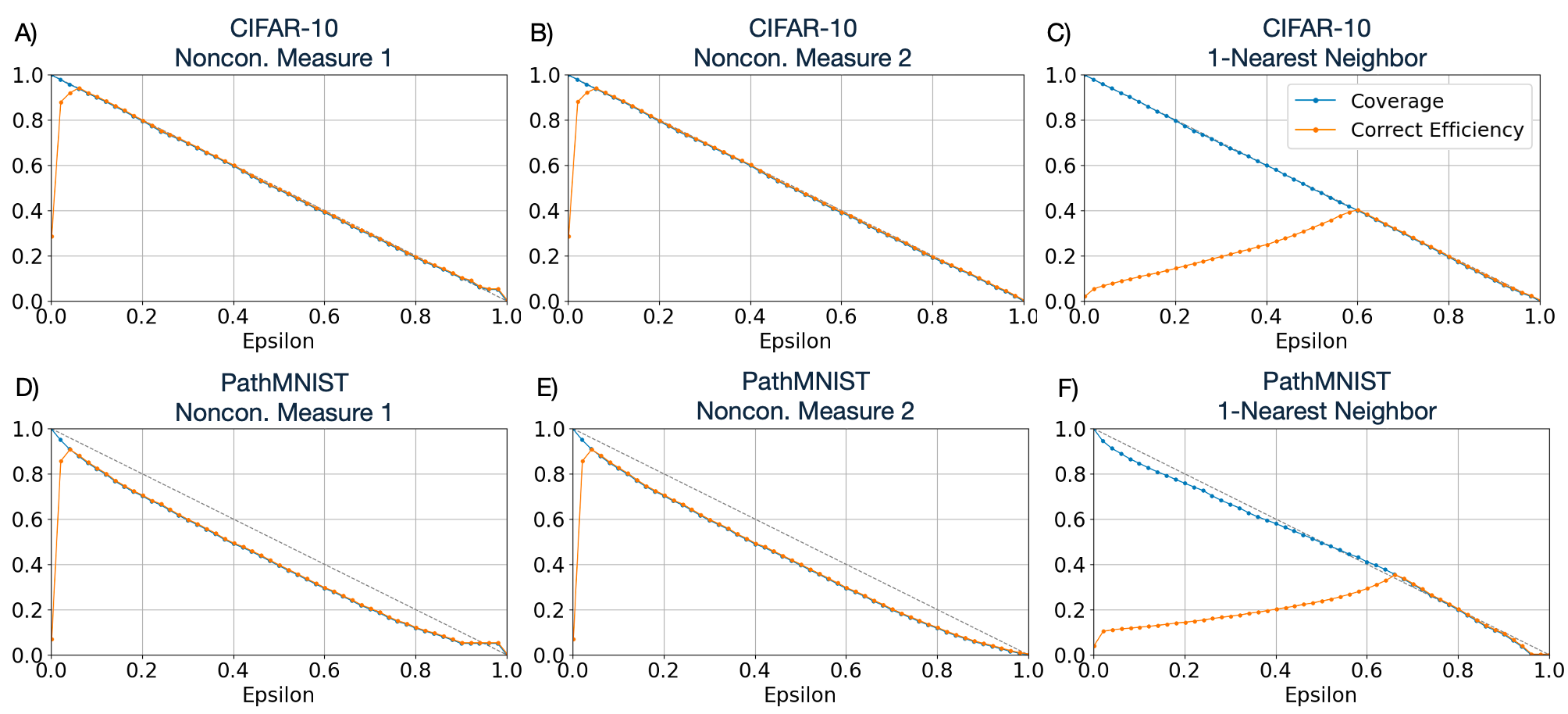}
    \vspace*{-6px}
    \caption{Coverage and correct efficiency curves when changing the significance level $\varepsilon$ for three prior conformal prediction methods. Hyperparameter used for (B), (E): $\gamma=1$.}
    \vspace*{-13px}
    \label{fig:coverage-prior}
\end{figure*}

In Fig.~\ref{fig:coverage} and Appendix~\ref{app:coverage-nosplit-all}, we analyze how coverage and correct efficiency change with different values of $\varepsilon$ in order to assess the validity of CONFINE. For CIFAR-10, when using outputs from the softmax layer as feature embeddings (Fig.~\ref{fig:coverage}A), the coverage curve is always above the diagonal line, confirming the validity of CONFINE. If we use the outputs from the layer before the fully-connected layer in ResNet18 ($l=50$), the coverage curve is very slightly below the diagonal line for some values of $\varepsilon$ (Fig.~\ref{fig:coverage}B), probably because the features extracted from this layer are not fully exchangeable. Since this disparity is small and ignorable, we say that CIFAR-10 is pseudo-exchangeable and that CONFINE is approximately valid in this case.

With PathMNIST, on the other hand, when using the second last layer ($l=50$) for feature extraction, the coverage curve appears to be much lower than the diagonal line (Fig.~\ref{fig:coverage}D), indicating that such a patient-based dataset is not exchangeable. Using the softmax layer pushes the coverage curve much higher, making it align with the diagonal line except for a few $\varepsilon$ values, suggesting approximate validity (Fig.~\ref{fig:coverage}C). This demonstrates that exchangeability depends on which layer we choose to extract features from and CONFINE's validity can extend across different types of datasets and settings, provided that an appropriate feature extraction layer is chosen.

Comparisons with the three prior conformal prediction methods show similar coverage outcomes (Fig.~\ref{fig:coverage-prior} and Appendix~\ref{app:prior-coverage-all}). CIFAR-10's coverage curves mostly align well with the diagonal line, suggesting the pseudo-exchangeability of CIFAR-10 and the approximate validity of these prior methods. For PathMNIST, the coverage curves are also far below the diagonal line, indicating the non-exchangeability of the dataset. After random shuffling and re-splitting into training, validation, and test sets, PathMNIST becomes exchangeable (Fig.~\ref{fig:pathmnist-shuffle}), which verifies that the non-exchangeability of PathMNIST is due to the distribution shift caused by patient-wise split.

When looking at the correct efficiency curves of CONFINE, after the first few very low $\varepsilon$ values, CONFINE's correct efficiency curves are identical to the coverage curves (Fig.~\ref{fig:coverage}). Since coverage is the fraction of prediction sets that contain the correct label and correct efficiency is the fraction of those that contain exactly the correct label, this overlap indicates that when $\varepsilon$ is not extremely small, all prediction sets that contain the correct label are of size 1, demonstrating CONFINE's precision and reliability. This trend is also present when using the two prior nonconformity measures (Eqs.~(\ref{noncon1}) and (\ref{noncon2})) in Fig.~\ref{fig:coverage-prior}. However, with the 1-Nearest Neighbor conformal predictor, the coverage and correct efficiency curves only overlap when $\varepsilon$ is very large (Fig.~\ref{fig:coverage-prior} C, F) because the prediction sets tend to contain multiple labels to enforce a high coverage allowed by $\varepsilon$. This indicates that 1-Nearest Neighbor is not as efficient and precise as CONFINE and the other prior methods.

\begin{figure}[ht]
  \centering
  \begin{minipage}[t]{0.29\textwidth}
    \includegraphics[width=1.05\textwidth]{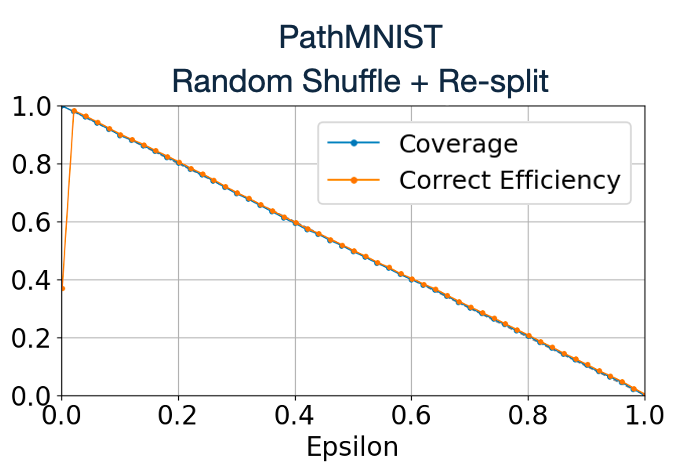}
    \vspace*{-15px}
  \end{minipage}
  \hfill
  \begin{minipage}[t]{0.68\textwidth}
    \raisebox{-0.2\baselineskip}{\includegraphics[width=\linewidth]{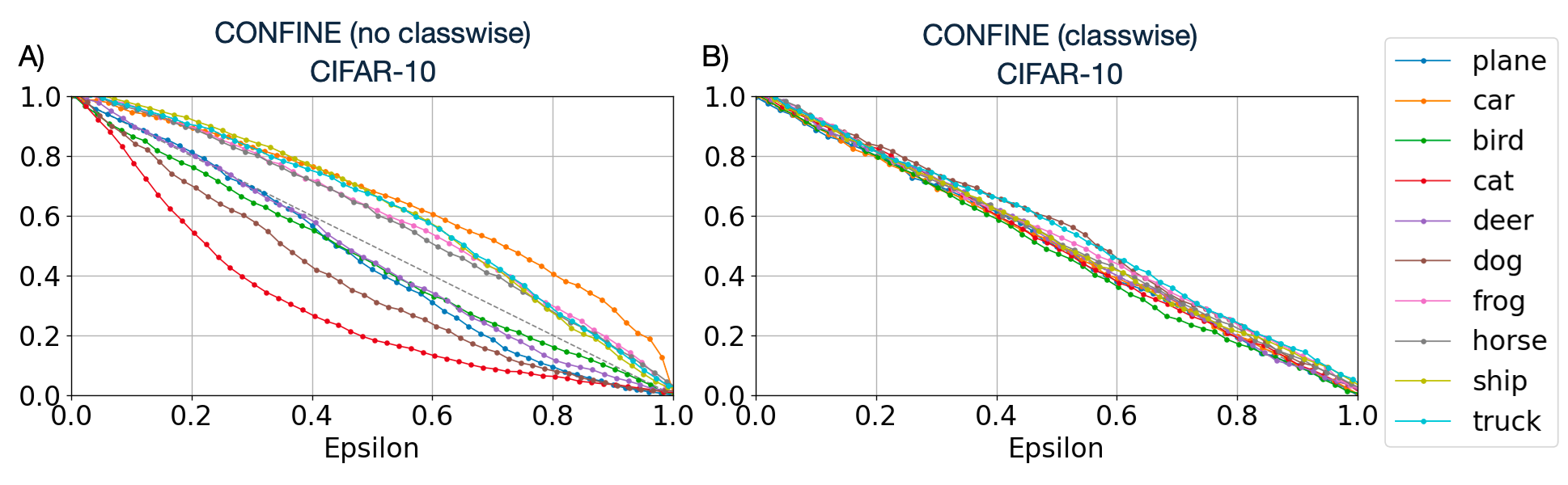}}
    \vspace*{-15px}
  \end{minipage}
  \vspace*{-15px}
\end{figure}
\begin{figure}[h]
  \centering
    \begin{minipage}[t]{0.3\textwidth}
    \caption{Coverage curve of PathMNIST after random shuffling and re-splitting suggests exchangeability.}
    \label{fig:pathmnist-shuffle}
  \end{minipage}
  \hspace{3px}
  \begin{minipage}[t]{0.67\textwidth}
    \caption{Class-wise coverage of CONFINE on CIFAR-10: (A) without and (B) with the class-wise split of the calibration set. With the class-wise split, class-conditional coverage is achieved. Hyperparameters used: $l=50,k=1$.}
    \label{fig:classwise-coverage}
  \end{minipage}
  \vspace*{-5px}
\end{figure}

\subsection{CONFINE Can Achieve Class-Conditional Coverage}
CONFINE not only ensures marginal coverage (Eq.~(\ref{marginal-coverage})) but can also achieve class-conditional coverage (Eq.~(\ref{cc-coverage})), a more stringent and informative measure of a conformal predictor's predictive power. Class-conditional coverage is particularly vital for applications, such as in healthcare settings, where fairness and balanced performance across all disease categories are crucial.

Without the class-wise split, CONFINE effectively achieves marginal coverage (Fig.~\ref{fig:coverage}B), but does not always provide satisfactory class-conditional coverage, as shown in Fig.~\ref{fig:classwise-coverage}A where only 4 out of 10 classes have coverages above the set significance level $1-\varepsilon$ for CIFAR-10. To address this, we introduce the class-wise split of the calibration set to ensure more tailored and accurate $p$-values for each individual class. After applying the class-wise split, CONFINE successfully achieves class-conditional coverage for all classes (Fig.~\ref{fig:classwise-coverage}B and Appendix~\ref{app:coverage-split-all}), significantly enhancing the model's utility and fairness. Although the class-wise split tends to very slightly reduce the overall accuracy and correct efficiency (Appendix~\ref{app:res-classwise}) due to the smaller data pool for each class-specific calibration split that involves more noise, it substantially increases the trustworthiness and applicability of CONFINE in diverse settings.

\section{Discussions and Limitations}\label{sec:limitations}
By providing nearest neighbor explanations and certainty estimates, CONFINE fosters a greater understanding of model behavior and highlights ambiguous cases, which the user can then further investigate. CONFINE's enhancement in accuracy, correct efficiency, overall coverage, and class-conditional coverage proves its reliability, precision, and statistical robustness. It is a promising approach for improving the interpretability and trustworthiness of neural network predictions to help the user make more informed decisions. However, an important limitation of CONFINE is its trade-off of computational efficiency for interpretability. CONFINE takes $O(dN_t)$ extra time and space due to the nearest neighbors calculation, where $d$ is the size of the extracted feature vector and $N_t$ is the size of the proper training set. A detailed analysis is included in Appendix~\ref{sec:comp-complexity}. In addition, extremely high confidence with CONFINE is not attainable because the prediction sets would encompass all classes when $\varepsilon$ is set to extremely low values. Also, if the exchangeability assumption is not satisfied, a strict coverage guarantee cannot be provided.

In future work, to reduce the computational burden of $k$-nearest neighbors, prototypes can be identified in the training set. For datasets known to be unexchangeable, such as patient-based medical datasets, modifications of conformal prediction can be employed to ensure validity \citep{barber2023conformal}.

\section{Conclusion}\label{sec:conclusion}
We presented a framework called CONFINE that applies conformal prediction to neural networks, leading to both interpretability and exceptional performance gains. By combining example-based explanations and statistically robust credibility and confidence quantifications, CONFINE not only enhances the transparency of the model's decision-making process but also boosts accuracy and consistently achieves higher correct efficiency than previous methods, highlighting its potential for settings where both accuracy and precision are critical. CONFINE is versatile and adaptable to any pre-trained machine learning classifier without any need for modification. Although CONFINE trades off computational costs for interpretability, it marks a significant advancement in improving the transparency and trustworthiness of deep learning models to enable broader utility in critical domains like healthcare.

\bibliographystyle{plainnat}
\bibliography{mybib}

\begin{thebibliography}{51}
\providecommand{\natexlab}[1]{#1}
\providecommand{\url}[1]{\texttt{#1}}
\expandafter\ifx\csname urlstyle\endcsname\relax
  \providecommand{\doi}[1]{doi: #1}\else
  \providecommand{\doi}{doi: \begingroup \urlstyle{rm}\Url}\fi

\bibitem[Abdar et~al.(2021)Abdar, Pourpanah, Hussain, Rezazadegan, Liu, Ghavamzadeh, Fieguth, Cao, Khosravi, Acharya, Makarenkov, and Nahavandi]{Abdar_Pourpanah_Hussain_Rezazadegan_Liu_Ghavamzadeh_Fieguth_Cao_Khosravi_Acharya_et_2021}
Moloud Abdar, Farhad Pourpanah, Sadiq Hussain, Dana Rezazadegan, Li~Liu, Mohammad Ghavamzadeh, Paul Fieguth, Xiaochun Cao, Abbas Khosravi, U.~Rajendra Acharya, Vladimir Makarenkov, and Saeid Nahavandi.
\newblock {A review of uncertainty quantification in deep learning: Techniques, applications and challenges}.
\newblock \emph{Information Fusion}, 76:\penalty0 243–297, December 2021.

\bibitem[Angelov and Soares(2020)]{angelov2019explainable}
Plamen Angelov and Eduardo Soares.
\newblock {Towards explainable deep neural networks (xDNN)}.
\newblock \emph{Neural Networks}, 130:\penalty0 185--194, 2020.

\bibitem[Ashukha et~al.(2020)Ashukha, Lyzhov, Molchanov, and Vetrov]{ashukha2021pitfalls}
Arsenii Ashukha, Alexander Lyzhov, Dmitry Molchanov, and Dmitry Vetrov.
\newblock {Pitfalls of in-domain uncertainty estimation and ensembling in deep learning}.
\newblock \emph{arXiv preprint arXiv:2002.06470}, 2020.

\bibitem[Bach et~al.(2015)Bach, Binder, Montavon, Klauschen, Müller, and Samek]{Bach_Binder_Montavon_Klauschen_Müller_Samek_2015}
Sebastian Bach, Alexander Binder, Grégoire Montavon, Frederick Klauschen, Klaus-Robert Müller, and Wojciech Samek.
\newblock On pixel-wise explanations for non-linear classifier decisions by layer-wise relevance propagation.
\newblock \emph{PLOS ONE}, 10\penalty0 (7):\penalty0 e0130140, July 2015.

\bibitem[Banerji et~al.(2023)Banerji, Chakraborti, Harbron, and MacArthur]{aitool}
Christopher R.~S. Banerji, Tapabrata Chakraborti, Chris Harbron, and Ben~D. MacArthur.
\newblock {Clinical AI tools must convey predictive uncertainty for each individual patient}.
\newblock \emph{Nature Medicine}, 29\penalty0 (12):\penalty0 2996--2998, 2023.

\bibitem[Barber et~al.(2023)Barber, Candes, Ramdas, and Tibshirani]{barber2023conformal}
Rina~Foygel Barber, Emmanuel~J. Candes, Aaditya Ramdas, and Ryan~J. Tibshirani.
\newblock {Conformal prediction beyond exchangeability}.
\newblock \emph{The Annals of Statistics}, 51\penalty0 (2):\penalty0 816--845, 2023.

\bibitem[Bilic et~al.(2023)Bilic, Christ, Li, Vorontsov, Ben-Cohen, Kaissis, Szeskin, Jacobs, Mamani, Chartrand, Lohöfer, Holch, Sommer, Hofmann, Hostettler, Lev-Cohain, Drozdzal, Amitai, Vivanti, Sosna, Ezhov, Sekuboyina, Navarro, Kofler, Paetzold, Shit, Hu, Lipková, Rempfler, Piraud, Kirschke, Wiestler, Zhang, Hülsemeyer, Beetz, Ettlinger, Antonelli, Bae, Bellver, Bi, Chen, Chlebus, Dam, Dou, Fu, Georgescu, Giró-i Nieto, Gruen, Han, Heng, Hesser, Moltz, Igel, Isensee, Jäger, Jia, Kaluva, Khened, Kim, Kim, Kim, Kohl, Konopczynski, Kori, Krishnamurthi, Li, Li, Li, Li, Lowengrub, Ma, Maier-Hein, Maninis, Meine, Merhof, Pai, Perslev, Petersen, Pont-Tuset, Qi, Qi, Rippel, Roth, Sarasua, Schenk, Shen, Torres, Wachinger, Wang, Weninger, Wu, Xu, Yang, Yu, Yuan, Yue, Zhang, Cardoso, Bakas, Braren, Heinemann, Pal, Tang, Kadoury, Soler, van Ginneken, Greenspan, Joskowicz, and Menze]{Bilic_Christ_Li_2023}
Patrick Bilic, Patrick Christ, Hongwei~Bran Li, Eugene Vorontsov, Avi Ben-Cohen, Georgios Kaissis, Adi Szeskin, Colin Jacobs, Gabriel Efrain~Humpire Mamani, Gabriel Chartrand, Fabian Lohöfer, Julian~Walter Holch, Wieland Sommer, Felix Hofmann, Alexandre Hostettler, Naama Lev-Cohain, Michal Drozdzal, Michal~Marianne Amitai, Refael Vivanti, Jacob Sosna, Ivan Ezhov, Anjany Sekuboyina, Fernando Navarro, Florian Kofler, Johannes~C. Paetzold, Suprosanna Shit, Xiaobin Hu, Jana Lipková, Markus Rempfler, Marie Piraud, Jan Kirschke, Benedikt Wiestler, Zhiheng Zhang, Christian Hülsemeyer, Marcel Beetz, Florian Ettlinger, Michela Antonelli, Woong Bae, Míriam Bellver, Lei Bi, Hao Chen, Grzegorz Chlebus, Erik~B. Dam, Qi~Dou, Chi-Wing Fu, Bogdan Georgescu, Xavier Giró-i Nieto, Felix Gruen, Xu~Han, Pheng-Ann Heng, Jürgen Hesser, Jan~Hendrik Moltz, Christian Igel, Fabian Isensee, Paul Jäger, Fucang Jia, Krishna~Chaitanya Kaluva, Mahendra Khened, Ildoo Kim, Jae-Hun Kim, Sungwoong Kim, Simon Kohl, Tomasz Konopczynski,
  Avinash Kori, Ganapathy Krishnamurthi, Fan Li, Hongchao Li, Junbo Li, Xiaomeng Li, John Lowengrub, Jun Ma, Klaus Maier-Hein, Kevis-Kokitsi Maninis, Hans Meine, Dorit Merhof, Akshay Pai, Mathias Perslev, Jens Petersen, Jordi Pont-Tuset, Jin Qi, Xiaojuan Qi, Oliver Rippel, Karsten Roth, Ignacio Sarasua, Andrea Schenk, Zengming Shen, Jordi Torres, Christian Wachinger, Chunliang Wang, Leon Weninger, Jianrong Wu, Daguang Xu, Xiaoping Yang, Simon Chun-Ho Yu, Yading Yuan, Miao Yue, Liping Zhang, Jorge Cardoso, Spyridon Bakas, Rickmer Braren, Volker Heinemann, Christopher Pal, An~Tang, Samuel Kadoury, Luc Soler, Bram van Ginneken, Hayit Greenspan, Leo Joskowicz, and Bjoern Menze.
\newblock {The Liver Tumor Segmentation Benchmark (LiTS)}.
\newblock \emph{Medical Image Analysis}, 84:\penalty0 102680, February 2023.

\bibitem[Chen et~al.(2019)Chen, Li, Tao, Barnett, Rudin, and Su]{chen2019looks}
Chaofan Chen, Oscar Li, Daniel Tao, Alina Barnett, Cynthia Rudin, and Jonathan~K. Su.
\newblock {This looks like that: Deep learning for interpretable image recognition}.
\newblock \emph{Advances in Neural Information Processing Systems}, 32, 2019.

\bibitem[Cobb and Jalaian(2021)]{cobb2020scaling}
Adam~D. Cobb and Brian Jalaian.
\newblock {Scaling Hamiltonian Monte Carlo inference for Bayesian neural networks with symmetric splitting}.
\newblock In \emph{Uncertainty in Artificial Intelligence}, pages 675--685. PMLR, 2021.

\bibitem[Deng et~al.(2009)Deng, Dong, Socher, Li, Li, and Fei-Fei]{Deng_Dong_Socher_Li_Li_Fei-Fei_2009}
Jia Deng, Wei Dong, Richard Socher, Li-Jia Li, Kai Li, and Li~Fei-Fei.
\newblock {ImageNet: A large-scale hierarchical image database}.
\newblock In \emph{2009 IEEE Conference on Computer Vision and Pattern Recognition}, page 248–255, June 2009.

\bibitem[Fong and Vedaldi(2017)]{Fong_2017}
Ruth~C. Fong and Andrea Vedaldi.
\newblock Interpretable explanations of black boxes by meaningful perturbation.
\newblock In \emph{IEEE International Conference on Computer Vision}. IEEE, October 2017.

\bibitem[Gal and Ghahramani(2015)]{gal2016bayesian}
Yarin Gal and Zoubin Ghahramani.
\newblock {Bayesian convolutional neural networks with Bernoulli approximate variational inference}.
\newblock \emph{arXiv preprint arXiv:1506.02158}, 2015.

\bibitem[Gal and Ghahramani(2016)]{pmlr-v48-gal16}
Yarin Gal and Zoubin Ghahramani.
\newblock {Dropout as a Bayesian approximation: Representing model uncertainty in deep learning}.
\newblock In Maria~Florina Balcan and Kilian~Q. Weinberger, editors, \emph{Proceedings of The 33rd International Conference on Machine Learning}, volume~48 of \emph{Proceedings of Machine Learning Research}, pages 1050--1059, New York, NY, USA, 20--22 Jun 2016. PMLR.

\bibitem[Hassantabar et~al.(2021)Hassantabar, Stefano, Ghanakota, Ferrari, Nicola, Bruno, Marino, Hamidouche, and Jha]{hassantabar2020coviddeep}
Shayan Hassantabar, Novati Stefano, Vishweshwar Ghanakota, Alessandra Ferrari, Gregory~N. Nicola, Raffaele Bruno, Ignazio~R. Marino, Kenza Hamidouche, and Niraj~K. Jha.
\newblock {CovidDeep: SARS-CoV-2/COVID-19 test based on wearable medical sensors and efficient neural networks}.
\newblock \emph{IEEE Transactions on Consumer Electronics}, 67\penalty0 (4):\penalty0 244--256, 2021.

\bibitem[He et~al.(2016)He, Zhang, Ren, and Sun]{he2015deep}
Kaiming He, Xiangyu Zhang, Shaoqing Ren, and Jian Sun.
\newblock {Deep residual learning for image recognition}.
\newblock In \emph{Proceedings of the IEEE Conference on Computer Vision and Pattern Recognition}, pages 770--778, 2016.

\bibitem[Held(2022{\natexlab{a}})]{hf_mnli}
Will Held.
\newblock {Hugging Face - WillHeld/roberta-base-mnli}, 2022{\natexlab{a}}.
\newblock URL \url{https://huggingface.co/WillHeld/roberta-base-mnli}.

\bibitem[Held(2022{\natexlab{b}})]{hf_sst2}
Will Held.
\newblock {Hugging Face - WillHeld/roberta-base-sst2}, 2022{\natexlab{b}}.
\newblock URL \url{https://huggingface.co/WillHeld/roberta-base-sst2}.

\bibitem[Hoffer and Ailon(2015)]{hoffer2018deep}
Elad Hoffer and Nir Ailon.
\newblock {Deep metric learning using triplet network}.
\newblock In \emph{Similarity-Based Pattern Recognition: Third International Workshop, SIMBAD, Proceedings 3}, pages 84--92. Springer, 2015.

\bibitem[Jetley et~al.(2018)Jetley, Lord, Lee, and Torr]{jetley2018learn}
Saumya Jetley, Nicholas~A. Lord, Namhoon Lee, and Philip H.~S. Torr.
\newblock {Learn to pay attention}.
\newblock \emph{arXiv preprint arXiv:1804.02391}, 2018.

\bibitem[Jospin et~al.(2022)Jospin, Laga, Boussaid, Buntine, and Bennamoun]{Jospin_2022}
Laurent~Valentin Jospin, Hamid Laga, Farid Boussaid, Wray Buntine, and Mohammed Bennamoun.
\newblock {Hands-on Bayesian neural networks—A tutorial for deep learning users}.
\newblock \emph{IEEE Computational Intelligence Magazine}, 17\penalty0 (2):\penalty0 29–48, May 2022.

\bibitem[Kather et~al.(2018)Kather, Halama, and Marx]{Kather_Halama_Marx_2018}
Jakob~Nikolas Kather, Niels Halama, and Alexander Marx.
\newblock {100,000 histological images of human colorectal cancer and healthy tissue}.
\newblock \emph{Zenodo}, April 2018.

\bibitem[Koh and Liang(2017)]{koh2020understanding}
Pang~Wei Koh and Percy Liang.
\newblock {Understanding black-box predictions via influence functions}.
\newblock In \emph{International Conference on Machine Learning}, pages 1885--1894. PMLR, 2017.

\bibitem[Krizhevsky(2009)]{Krizhevsky09learningmultiple}
Alex Krizhevsky.
\newblock {Learning multiple layers of features from tiny images}.
\newblock Technical report, 2009.

\bibitem[Kwon et~al.(2020)Kwon, Won, Kim, and Paik]{Kwon_Won_Kim_Paik_2020}
Yongchan Kwon, Joong-Ho Won, Beom~Joon Kim, and Myunghee~Cho Paik.
\newblock {Uncertainty quantification using Bayesian neural networks in classification: Application to biomedical image segmentation}.
\newblock \emph{Computational Statistics \& Data Analysis}, 142:\penalty0 106816, February 2020.

\bibitem[Lakshminarayanan et~al.(2017)Lakshminarayanan, Pritzel, and Blundell]{Lakshminarayanan_Pritzel_Blundell_2017}
Balaji Lakshminarayanan, Alexander Pritzel, and Charles Blundell.
\newblock Simple and scalable predictive uncertainty estimation using deep ensembles.
\newblock In I.~Guyon, U.~Von Luxburg, S.~Bengio, H.~Wallach, R.~Fergus, S.~Vishwanathan, and R.~Garnett, editors, \emph{Advances in Neural Information Processing Systems}, volume~30. Curran Associates, Inc., 2017.

\bibitem[Lee et~al.(2019)Lee, Wang, Vlahov, Brar, and Theodorou]{lee2020ensemble}
Keuntaek Lee, Ziyi Wang, Bogdan Vlahov, Harleen Brar, and Evangelos~A. Theodorou.
\newblock {Ensemble Bayesian decision making with redundant deep perceptual control policies}.
\newblock In \emph{IEEE International Conference On Machine Learning And Applications}, pages 831--837. IEEE, 2019.

\bibitem[Liu et~al.(2019)Liu, Ott, Goyal, Du, Joshi, Chen, Levy, Lewis, Zettlemoyer, and Stoyanov]{liu2019roberta}
Yinhan Liu, Myle Ott, Naman Goyal, Jingfei Du, Mandar Joshi, Danqi Chen, Omer Levy, Mike Lewis, Luke Zettlemoyer, and Veselin Stoyanov.
\newblock {Roberta: A robustly optimized bert pretraining approach}.
\newblock \emph{arXiv preprint arXiv:1907.11692}, 2019.

\bibitem[Lundberg and Lee(2017)]{lundberg2017unified}
Scott~M. Lundberg and Su-In Lee.
\newblock {A unified approach to interpreting model predictions}.
\newblock \emph{Advances in Neural Information Processing Systems}, 30, 2017.

\bibitem[Neal(1996)]{Neal_1996}
Radford~M. Neal.
\newblock \emph{{Bayesian Learning for Neural Networks}}, volume 118 of \emph{Lecture Notes in Statistics}.
\newblock Springer, NY, 1996.

\bibitem[Nouretdinov et~al.(2001)Nouretdinov, Vovk, Vyugin, and Gammerman]{Nouretdinov_Vovk_Vyugin_Gammerman_2001}
Ilia Nouretdinov, Volodya Vovk, Michael Vyugin, and Alex Gammerman.
\newblock Pattern recognition and density estimation under the general i.i.d. assumption.
\newblock In David Helmbold and Bob Williamson, editors, \emph{Computational Learning Theory}, page 337–353, Berlin, Heidelberg, 2001. Springer.

\bibitem[Papadopoulos(2008)]{Papadopoulos_2008}
Harris Papadopoulos.
\newblock \emph{Inductive Conformal Prediction: Theory and Application to Neural Networks}.
\newblock InTech, August 2008.

\bibitem[Papadopoulos et~al.(2007)Papadopoulos, Vovk, and Gammerman]{confnn}
Harris Papadopoulos, Volodya Vovk, and Alex Gammerman.
\newblock Conformal prediction with neural networks.
\newblock In \emph{IEEE International Conference on Tools with Artificial Intelligence}, volume~2, pages 388--395, 2007.

\bibitem[Papernot and McDaniel(2018)]{papernot2018deep}
Nicolas Papernot and Patrick McDaniel.
\newblock {Deep k-nearest neighbors: Towards confident, interpretable and robust deep learning}.
\newblock \emph{arXiv preprint arXiv:1803.04765}, 2018.

\bibitem[Ribeiro et~al.(2016)Ribeiro, Singh, and Guestrin]{ribeiro2016modelagnostic}
Marco~Tulio Ribeiro, Sameer Singh, and Carlos Guestrin.
\newblock {Model-agnostic interpretability of machine learning}.
\newblock \emph{arXiv preprint arXiv:1606.05386}, 2016.

\bibitem[Selvaraju et~al.(2019)Selvaraju, Cogswell, Das, Vedantam, Parikh, and Batra]{Selvaraju_2019}
Ramprasaath~R. Selvaraju, Michael Cogswell, Abhishek Das, Ramakrishna Vedantam, Devi Parikh, and Dhruv Batra.
\newblock {Grad-CAM: Visual explanations from deep networks via gradient-based localization}.
\newblock \emph{International Journal of Computer Vision}, 128\penalty0 (2):\penalty0 336–359, October 2019.

\bibitem[Sensoy et~al.(2018)Sensoy, Kaplan, and Kandemir]{sensoy2018evidential}
Murat Sensoy, Lance Kaplan, and Melih Kandemir.
\newblock {Evidential deep learning to quantify classification uncertainty}.
\newblock \emph{Advances in Neural Information Processing Systems}, 31, 2018.

\bibitem[Shafer and Vovk(2008)]{shafer2007tutorial}
Glenn Shafer and Vladimir Vovk.
\newblock {A tutorial on conformal prediction}.
\newblock \emph{Journal of Machine Learning Research}, 9\penalty0 (3), 2008.

\bibitem[Shrikumar et~al.(2017)Shrikumar, Greenside, and Kundaje]{shrikumar2019learning}
Avanti Shrikumar, Peyton Greenside, and Anshul Kundaje.
\newblock {Learning important features through propagating activation differences}.
\newblock In \emph{International Conference on Machine Learning}, pages 3145--3153. PMLR, 2017.

\bibitem[Silva et~al.(2018)Silva, Fernandes, Cardoso, and Cardoso]{Silva_Fernandes_Cardoso_Cardoso_2018}
Wilson Silva, Kelwin Fernandes, Maria~J. Cardoso, and Jaime~S. Cardoso.
\newblock \emph{{Towards Complementary Explanations Using Deep Neural Networks}}, volume 11038 of \emph{Lecture Notes in Computer Science}, page 133–140.
\newblock Springer International Publishing, Cham, 2018.

\bibitem[Springenberg et~al.(2014)Springenberg, Dosovitskiy, Brox, and Riedmiller]{springenberg2015striving}
Jost~Tobias Springenberg, Alexey Dosovitskiy, Thomas Brox, and Martin Riedmiller.
\newblock {Striving for simplicity: The all convolutional net}.
\newblock \emph{arXiv preprint arXiv:1412.6806}, 2014.

\bibitem[Szegedy et~al.(2013)Szegedy, Zaremba, Sutskever, Bruna, Erhan, Goodfellow, and Fergus]{szegedy2014intriguing}
Christian Szegedy, Wojciech Zaremba, Ilya Sutskever, Joan Bruna, Dumitru Erhan, Ian Goodfellow, and Rob Fergus.
\newblock {Intriguing properties of neural networks}.
\newblock \emph{arXiv preprint arXiv:1312.6199}, 2013.

\bibitem[van~der Velden et~al.(2022)van~der Velden, Kuijf, Gilhuijs, and Viergever]{Viergever_2022}
Bas H.~M. van~der Velden, Hugo~J. Kuijf, Kenneth G.~A. Gilhuijs, and Max~A. Viergever.
\newblock {Explainable artificial intelligence (XAI) in deep learning-based medical image analysis}.
\newblock \emph{Medical Image Analysis}, 79:\penalty0 102470, July 2022.

\bibitem[Vovk(2012)]{Vovk_2012}
Vladimir Vovk.
\newblock Conditional validity of inductive conformal predictors.
\newblock In \emph{Proceedings of the Asian Conference on Machine Learning}, page 475–490. PMLR, November 2012.

\bibitem[Vovk et~al.(2005)Vovk, Gammerman, and Shafer]{vovk2005}
Vladimir Vovk, Alex Gammerman, and Glenn Shafer.
\newblock \emph{{Algorithmic Learning in a Random World}}.
\newblock Springer-Verlag, Berlin, Heidelberg, 2005.

\bibitem[Wang et~al.(2018)Wang, Singh, Michael, Hill, Levy, and Bowman]{wang2019glue}
Alex Wang, Amanpreet Singh, Julian Michael, Felix Hill, Omer Levy, and Samuel~R. Bowman.
\newblock {GLUE: A multi-task benchmark and analysis platform for natural language understanding}.
\newblock \emph{arXiv preprint arXiv:1804.07461}, 2018.

\bibitem[Yang et~al.(2023)Yang, Shi, Wei, Liu, Zhao, Ke, Pfister, and Ni]{Yang_2023}
Jiancheng Yang, Rui Shi, Donglai Wei, Zequan Liu, Lin Zhao, Bilian Ke, Hanspeter Pfister, and Bingbing Ni.
\newblock {MedMNIST v2 - A large-scale lightweight benchmark for 2D and 3D biomedical image classification}.
\newblock \emph{Scientific Data}, 10\penalty0 (1), January 2023.

\bibitem[Zeiler and Fergus(2014)]{zeiler2013visualizing}
Matthew~D. Zeiler and Rob Fergus.
\newblock {Visualizing and understanding convolutional networks}.
\newblock In \emph{European Conference on Computer Vision, Part I 13}, pages 818--833. Springer, 2014.

\bibitem[Zhang et~al.(2020)Zhang, Kailkhura, and Han]{zhang2020mixnmatch}
Jize Zhang, Bhavya Kailkhura, and T.~Yong-Jin Han.
\newblock {Mix-n-match: Ensemble and compositional methods for uncertainty calibration in deep learning}.
\newblock In \emph{International Conference on Machine Learning}, pages 11117--11128. PMLR, 2020.

\bibitem[Zhou et~al.(2016)Zhou, Khosla, Lapedriza, Oliva, and Torralba]{zhou2015learning}
Bolei Zhou, Aditya Khosla, Agata Lapedriza, Aude Oliva, and Antonio Torralba.
\newblock {Learning deep features for discriminative localization}.
\newblock In \emph{Proceedings of the IEEE Conference on Computer Vision and Pattern Recognition}, pages 2921--2929, 2016.

\bibitem[Zintgraf et~al.(2017)Zintgraf, Cohen, Adel, and Welling]{zintgraf2017visualizing}
Luisa~M. Zintgraf, Taco~S. Cohen, Tameem Adel, and Max Welling.
\newblock {Visualizing deep neural network decisions: Prediction difference analysis}.
\newblock \emph{arXiv preprint arXiv:1702.04595}, 2017.

\bibitem[Zou et~al.(2023)Zou, Chen, Yuan, Shen, Wang, and Fu]{Zou_Chen_Yuan_Shen_Wang_Fu_2023}
Ke~Zou, Zhihao Chen, Xuedong Yuan, Xiaojing Shen, Meng Wang, and Huazhu Fu.
\newblock {A review of uncertainty estimation and its application in medical imaging}.
\newblock \emph{Meta-Radiology}, 1\penalty0 (1):\penalty0 100003, June 2023.

\end{thebibliography}


\newpage
\appendix
\begin{appendices}
\counterwithin{figure}{section}
\counterwithin{table}{section}

\section{Experimental Setup Details}\label{app:experiments}

\subsection{Dataset and Training Details}\label{app:tasks}
\paragraph{Medical Sensor Data Diagnosis (CovidDeep).} CovidDeep (under CC BY-NC-ND 4.0 License) \citep{hassantabar2020coviddeep} contains medical sensor data collected from 87 patients, categorized into healthy, COVID-symptomatic, and COVID-asymptomatic classes. The data types from smartwatch sensors include Galvanic skin response, skin temperature, inter-beat interval, oxygen saturation, systolic blood pressure, and diastolic blood pressure. We follow the original article's data pre-processing pipeline: the time-series data are segmented into windows, concatenated, and divided into 10,191 training, 3,947 validation, and 3,256 test samples based on a patient-wise split. Galvanic skin response (GSR), pulse oximeter (Ox), blood pressure (BP), and questionnaire (Q) are chosen through feature selection for the highest accuracy and concatenated into vectors of length 74. We use the training set as the proper training set and the validation set as the calibration set in CONFINE.

A pre-trained four-layer fully-connected deep neural network from the original article is used in CONFINE for CovidDeep. The network has an input size of 74, a first hidden dimension size of 256, second and third hidden dimension sizes of 128, and an output size of 3. We perform a hyperparameter grid search for the numbers of nearest neighbors $k=1,5,10,20,40,60$, layer indices $l = 0,1,2,3,\text{softmax}$, and for $l=\text{softmax}$, temperatures $T=0.0001,0.001,0.01,0.1,1,10$. Running CONFINE for CovidDeep on an AMD EPYC 7413 CPU takes around one minute and 550 MB of CPU memory for layers used in Fig.~\ref{fig:coverage-nosplit-all}.

\paragraph{Image Classification (CIFAR-10 and MedMNIST).} CIFAR-10 (under CC-BY-SA License) \citep{Krizhevsky09learningmultiple} contains 60,000 32$\times$32 images categorized into 10 classes: airplanes, cars, birds, cats, deer, dogs, frogs, horses, ships, and trucks. Among them, the 50,000 training images are divided into proper training and calibration sets in a 7:3 ratio, and the test set of 10,000 images is unchanged.

We use PathMNIST and OrganAMNIST datasets from MedMNIST v2 (under CC-BY 4.0 License) \citep{Yang_2023}. PathMNIST is aggregated from \citet{Kather_Halama_Marx_2018} and contains 100,000 28$\times$28 colon pathology image patches divided into 89,996 training and 10,004 validation images, and 7,180 additional test images obtained from a different clinical center. The images are categorized into nine classes: adipose, background, debris, lymphocytes, mucus, smooth muscle, normal colon mucosa, cancer-associated stroma, and colorectal adenocarcinoma epithelium. OrganAMNIST is aggregated from \citet{Bilic_Christ_Li_2023} and contains 58,830 28$\times$28 abdominal CT images divided into 34,561 training, 6,491 validation, and 17,778 test samples splitted based on source CT scans. The images are categorized into 11 classes: bladder, femur-left, femur-right, heart, kidney-left, kidney-right, liver, lung-left, lung-right, pancreas, and spleen. Again, we use the training set as the proper training set and the validation set as the calibration set in CONFINE.

For the image classification tasks, ResNet18 (under Apache License 2.0) \citep{he2015deep} pre-trained on ImageNet \citep{Deng_Dong_Socher_Li_Li_Fei-Fei_2009} is fine-tuned on the training and validation sets of CIFAR-10, PathMNIST, and OrganAMNIST, respectively. The images are all randomly transformed, normalized, and resized to 224$\times$224 before being fed to the network. During fine-tuning, we use a batch size of 128, a learning rate of 0.0001, and the Adam optimizer. We pick the model with the highest validation accuracy and use that as the pre-trained network in CONFINE.  For CONFINE, we perform hyperparameter search for the numbers of nearest neighbors $k=1,5,10,20,40,50,60$, layer indices $l = 41,42,43,44,45,46,47,48,49,50,51,\text{softmax}$, where layer 50 is the layer before the fully-connected classification layer, and for $l=\text{softmax}$, temperatures $T=0.01, 0.1, 1, 10, 20, 40$.

Fine-tuning ResNet18 on an Nvidia A100 GPU takes about 80 minutes, 2.5 GB of CPU memory, and 7 GB of GPU memory. CONFINE is run on an AMD EPYC 7413 CPU with network inferences on an Nvidia A100 GPU. The exact time and memory consumption depends on the layer chosen. For layers used in Fig.~\ref{fig:coverage-nosplit-all}, for CIFAR-10, it takes about 13 minutes, 55 GB of CPU memory, and 7 GB of GPU memory; for PathMNIST, it takes about 18 minutes, 88 GB of CPU memory, and 12 GB of GPU memory; for OrganAMNIST, it takes about 20 minutes, 66 GB of CPU memory, and 12 GB of GPU memory. 

\paragraph{Natural Language Understanding (GLUE).} We use SST-2 and MNLI datasets of the GLUE benchmark (under CC-BY-SA License) \citep{wang2019glue} for evaluation. The sentiment analysis task, SST-2, aims to distinguish between negative and positive sentiments, given a sentence. It contains 67,349 training, 872 validation, and 1,821 test samples. In the textual entailment task, MNLI, given a premise sentence and a hypothesis sentence, the task aims to predict whether the premise entails the hypothesis, contradicts the hypothesis, or neither. We use MNLI-mismatched, which is the cross-domain textual entailment task, that contains 392,702 training, 9,832 validation, and 9,847 testing samples. For both datasets, we disregard the test sets since they are unlabeled. We use the validation samples as the test set in CONFINE. We split the training samples into the proper training and the calibration sets in a 7:3 ratio. For the pre-trained neural network, we use RoBERTa-base \citep{liu2019roberta} fine-tuned on each of these two tasks with their respective tokenizers (under MIT License) \citep{hf_mnli, hf_sst2}. For CONFINE, we use a batch size of 32 and perform hyperparameter search for the numbers of nearest neighbors $k=1,5,10,20,40,50,60$, layer indices $l = 145,146,147,148,149,150,151,\text{softmax}$, where $l=147$ is the layer before the classification head, and for $l=\text{softmax}$, temperatures $T=0.01, 0.1, 1, 5, 10, 20$.

CONFINE is run on an AMD EPYC 7413 CPU with network inferences on an Nvidia A100 GPU. The exact time and memory consumption depends on the layer chosen. For layers used in Fig.~\ref{fig:coverage-nosplit-all}, for SST-2, it takes about 5 minutes, 18 GB of CPU memory, and 6 GB of GPU memory; for MNLI, it takes about 80 minutes, 50 GB of CPU memory, and 24 GB of GPU memory.

\begin{table*}[t]
\caption{Hyperparameters used for results in Table~\ref{tab:accuracy} and Table~\ref{tab:accuracy-classwise}.}
\label{tab:accuracy-hparam}
\vskip 0.15in
\begin{center}
\footnotesize
\begin{tabularx}{\linewidth}{@{\extracolsep{\fill}} llc}
\toprule
Dataset & Method & Hyperparameters \\
\midrule
\multirow{5}{*}{CIFAR-10} & Noncon. Measure 2 & $\gamma=0.001, 0.01$ \\
 & \textbf{CONFINE-A} & $l=50,k=5$\\
 & \textbf{CONFINE-C} & $l=\text{softmax}, T=0.01, k=5,10,20$\\
 & \textbf{CONFINE-classwise-A} & $l=50,k=40$\\
 & \textbf{CONFINE-classwise-C} & $l=\text{softmax}, T=0.01, k=1$\\
\midrule
\multirow{5}{*}{PathMNIST} & Noncon. Measure 2 & $\gamma=0.001,0.01,0.1,1,10$\\
 & \textbf{CONFINE-A} & $l=50, k=20$ \\
 & \textbf{CONFINE-C} & $l=50, k=50$\\
 & \textbf{CONFINE-classwise-A} & $l=50,k=60$\\
 & \textbf{CONFINE-classwise-C} & $l=47,k=5$\\
\midrule
\multirow{5}{*}{OrganAMNIST} & Noncon. Measure 2 & $\gamma=0.001,0.01,0.1,1,10$\\
 & \textbf{CONFINE-A} & $l=49, k=5$ \\
 & \textbf{CONFINE-C} & $l=50, k=10$\\
 & \textbf{CONFINE-classwise-A} & $l=49,k=5$\\
 & \textbf{CONFINE-classwise-C} & $l=50,k=50$\\
 \midrule
\multirow{6}{*}{CovidDeep} & D$k$NN-A &  $k=60$ \\
 & D$k$NN-C & $k=3,5$ \\
 & Noncon. Measure 2 &  $\gamma=0.001,0.01, 0.1, 1$\\
 & \textbf{CONFINE-A/-C} & \begin{tabular}{@{}c@{}}$l=\text{softmax},T=0.0001,0.001,0.01$\\$k=1,5,10,20,40,60$\end{tabular}\\
 & \textbf{CONFINE-classwise-A/-C} & $l=\text{softmax},T=0.0001,k=1,5,10,20,40,60$\\
\midrule
\multirow{5}{*}{SST-2} & Noncon. Measure 2 & $\gamma=0.001,0.01,0.1,1,10$\\
 & \textbf{CONFINE-A} & $l=147,149,k=1$ \\
 & \textbf{CONFINE-C} & $l=149, k=5$ \\
 & \textbf{CONFINE-classwise-A} & $l=147,k=1$\\
 & \textbf{CONFINE-classwise-C} & $l=151,k=20,40,50,60$\\
\midrule
\multirow{5}{*}{MNLI} & Noncon. Measure 2 & $\gamma=10$\\
 & \textbf{CONFINE-A} & $l=147, k=20$ \\
 & \textbf{CONFINE-C} & $l=\text{softmax}, T=20, k=50$ \\
 & \textbf{CONFINE-classwise-A} & $l=147, k=60$ \\
 & \textbf{CONFINE-classwise-C} & $l=146, k=5$ \\
\bottomrule
\end{tabularx}
\end{center}
\vskip -0.1in
\end{table*}

\begin{figure}[ht]
    \centering
    \includegraphics[width=0.8\linewidth]{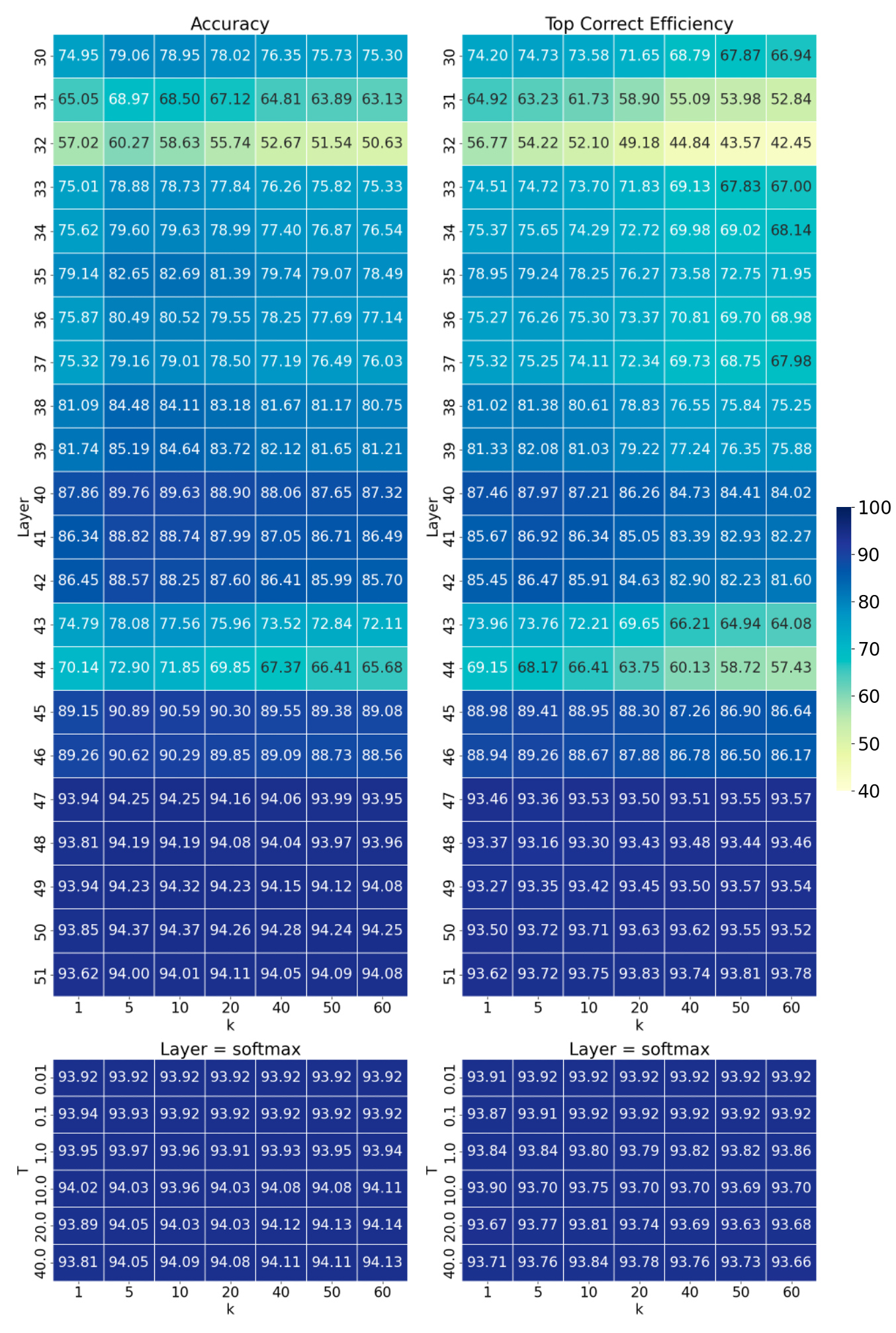}
    \vspace{-5px}
    \caption{Accuracy (left) and top correct efficiency (right) heatmaps from the hyperparameter search for CIFAR-10. Layer: the feature extraction layer $l$, $k$: the number of nearest neighbors, $T$: the temperature used for the softmax layer.}
    \label{fig:hparam-search}
    \vspace{-5px}
\end{figure}

\subsection{Hyperparameter Search}\label{app:hparams}

We report the highest accuracy and highest top correct efficiency results returned by hyperparameter searches in Tables~\ref{tab:accuracy} and \ref{tab:accuracy-classwise}. Detailed hyperparameters are listed in Table~\ref{tab:accuracy-hparam}.

Fig.~\ref{fig:hparam-search} illustrates the accuracy and top correct efficiency performances while changing the hyperparameters for CIFAR-10. Changing the feature extraction layer $l$ imposes much larger changes in performance than changing the number of nearest neighbors $k$ and the softmax temperature $T$. Generally, $k=5,10,20$ have the highest prediction accuracies and $k=1,5,10$ have the highest top correct efficiencies across each layer. When using the softmax layer, the accuracies and correct efficiencies only have very small differences among the different choices of $k$ and $T$. 

The layer indices in Fig.~\ref{fig:hparam-search} correspond to the layers in ResNet18, the pre-trained network for CIFAR-10 used in CONFINE. ResNet18 consists of a conv1 layer, followed by conv2, conv3, conv4, and conv5 layers each containing two 2-layer building blocks, and an average pooling and a fully connected layer at the end \citep{he2015deep}. Here $l=51$ is the final fully-connected layer and $l=50$ is the average pooling layer. $l=45$ to $49$ are the layers in the second building block of conv5, namely Conv2d, BatchNorm2d, ReLU, Conv2d, and BatchNorm2d layers in order. $l=38$ to $44$ are from the first building block of conv5, namely Conv2d, BatchNorm2d, ReLU, Conv2d, BatchNorm2d, and additional Conv2d and BatchNorm2d layers for downsampling. Similarly, $l=33$ to $37$ are the second building block of conv4, $l=26$ to $32$ are the first building block of conv4, and so on. From Fig.~\ref{fig:hparam-search}, we can see that the softmax layer and layers 51 and 50 perform best, with decreasing performance in each building block as we traverse the network upwards. Within each building block, the third to fifth layers (ReLU, Conv2d, and BatchNorm2d) perform better than the first two. The downsampling layers (layers 43, 44, 31, and 32) have the worst performance due to loss of information. A similar pattern is seen in all building blocks.

\section{Additional Results}

\begin{figure}
    \centering
    \includegraphics[width=\linewidth]{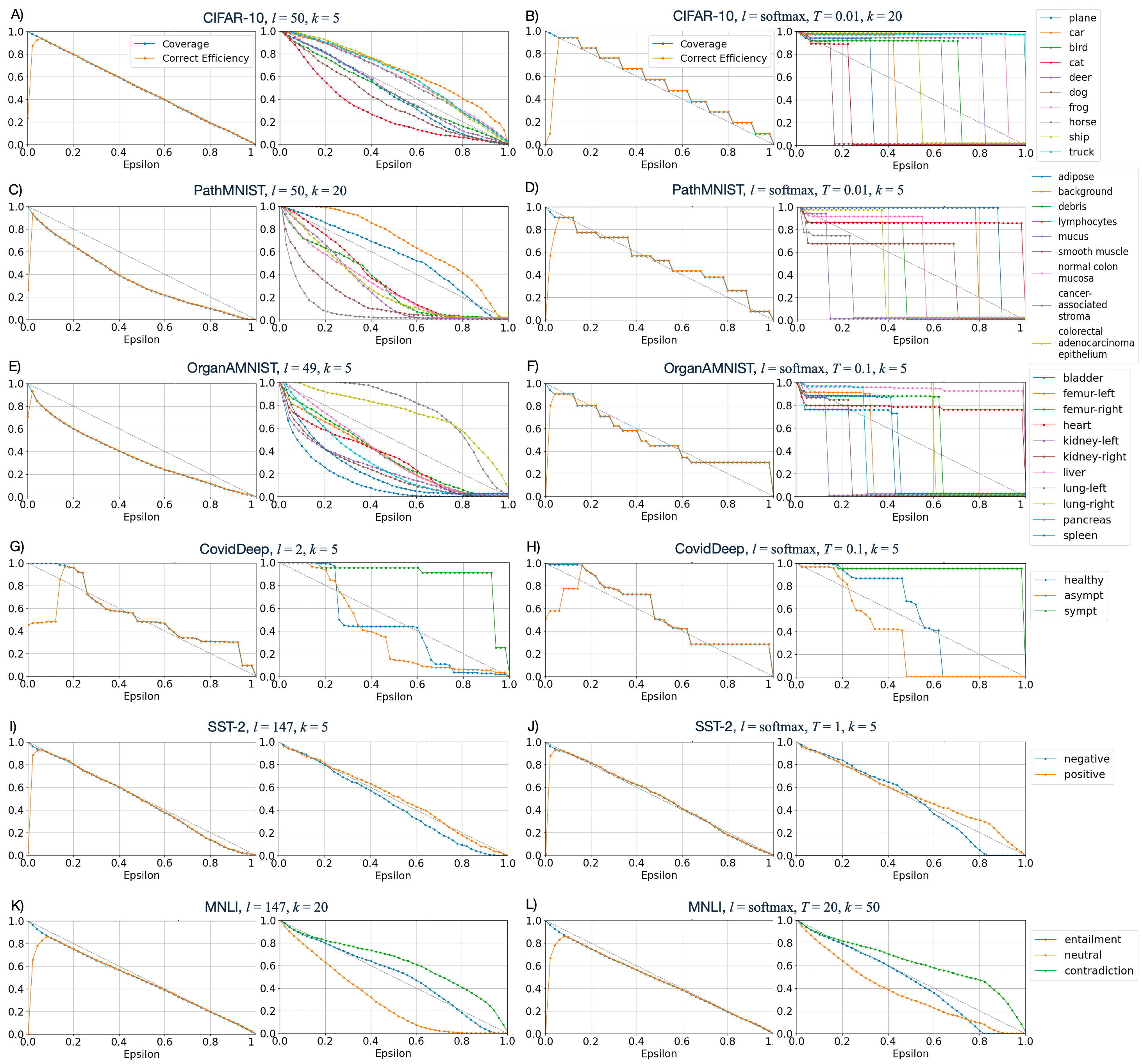}
    \caption{Overall coverage (left) and class-wise coverage (right) curves without the class-wise split of the calibration set for all six datasets using different hyperparameters.}
    \label{fig:coverage-nosplit-all}
\end{figure}

\subsection{Coverage Curves of CONFINE Without Class-wise Split}\label{app:coverage-nosplit-all}
Fig.~\ref{fig:coverage-nosplit-all} illustrates the overall coverage and correct efficiency curves and class-wise coverage curves for all six datasets without the class-wise split of the calibration set. Plots for a layer before the classification head that achieves high accuracy and the softmax layer are shown. Other layers before the classification head with different hyperparameters have results that look similar to Fig.~\ref{fig:coverage-nosplit-all} A, C, E, G, I, K, and the softmax layer with other hyperparameters have results that look similar to Fig.~\ref{fig:coverage-nosplit-all} B, D, F, H, J, L.

The overall coverage curves of CIFAR-10, CovidDeep, SST-2, and MNLI align well with the diagonal line. This good marginal coverage indicates the approximate exchangeability of the datasets and approximate validity of CONFINE (Fig.~\ref{fig:coverage-nosplit-all} A, B, G, H, I, J, K, L). For patient-based datasets, PathMNIST and OrganAMNIST, their overall coverage curves using a layer before the classification head are much lower than the diagonal line, indicating their non-exchangeability (Fig.~\ref{fig:coverage-nosplit-all} C, E). Nevertheless, approximate validity can be achieved by using the softmax layer in CONFINE (Fig.~\ref{fig:coverage-nosplit-all} D, F). Note that CovidDeep, despite also being a patient-based dataset, is roughly exchangeable, possibly because the distribution shift between training and testing samples is small (Fig.~\ref{fig:coverage-nosplit-all} G, H).

Without the class-wise split of the calibration set, the class-wise coverage curves show unsatisfactory results. Generally, less than half of the classes have coverage curves above the diagonal line, while others reside lower than the diagonal line, indicating failure to achieve class-conditional coverage.

\begin{figure}[tp]
    \centering
    \includegraphics[width=0.85\linewidth]{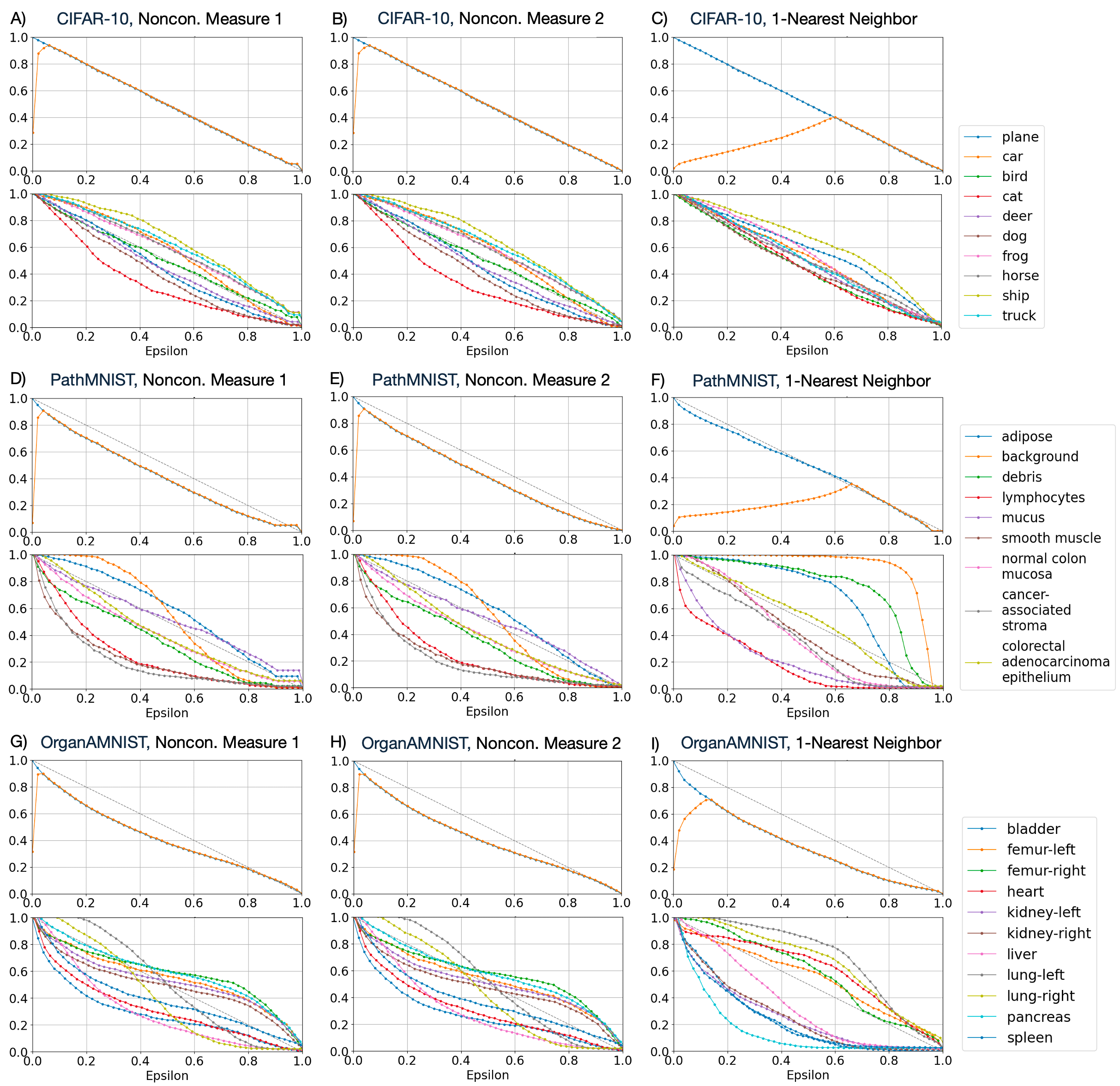}
    \caption{Overall coverage (top) and class-wise coverage (bottom) curves of prior conformal prediction methods for CIFAR-10, PathMNIST, and OrganAMNIST: Nonconformity Measure 1 (Eq.~(\ref{noncon1})), Nonconformity Measure 2 (Eq.~(\ref{noncon2})), and 1-Nearest Neighbor (Eq.~(\ref{1nn})). Hyperparameters used for Nonconformity Measure 2: $\gamma=1$.}
    \label{fig:prior-coverage-all-1}
\end{figure}

\begin{figure}[tp]
    \centering
    \includegraphics[width=\linewidth]{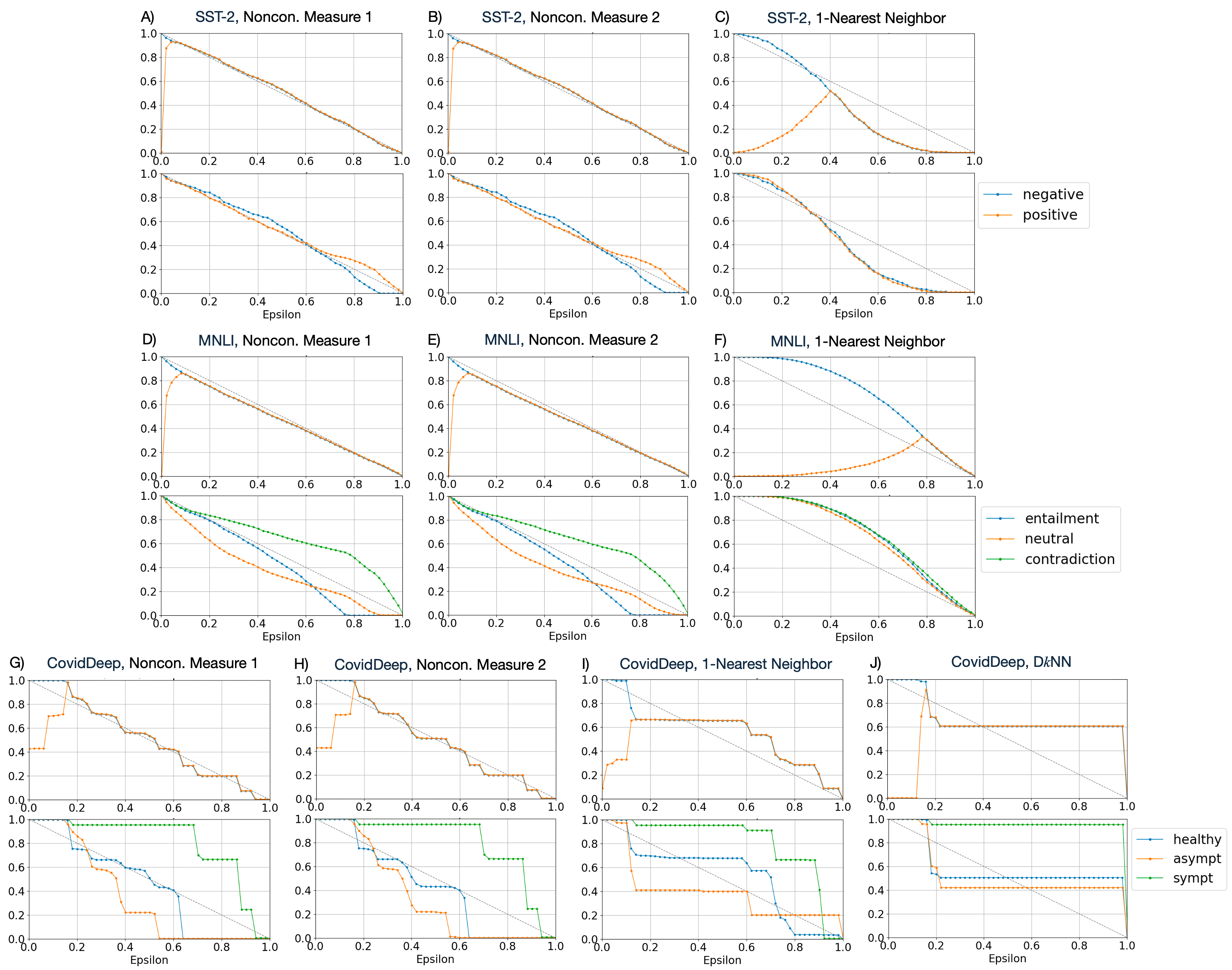}
    \caption{Overall coverage (top) and class-wise coverage (bottom) curves of prior conformal prediction methods for SST-2, MNLI, and CovidDeep: Nonconformity Measure 1 (Eq.~(\ref{noncon1})), Nonconformity Measure 2 (Eq.~(\ref{noncon2})), 1-Nearest Neighbor (Eq.~(\ref{1nn})), and D$k$NN \citep{papernot2018deep}. Hyperparameters used: Nonconformity Measure 2: $\gamma=1$, D$k$NN: $k=60$.}
    \label{fig:prior-coverage-all-2}
\end{figure}

\subsection{Coverage Curves of Prior Methods}\label{app:prior-coverage-all}

Overall coverage and correct efficiency curves and class-wise coverage curves of the prior conformal prediction methods in Figs.~\ref{fig:prior-coverage-all-1} and \ref{fig:prior-coverage-all-2} also verify the conclusions that PathMNIST and OrganAMNIST are unexchangeable and the others are exchangeable: the overall coverage curves of PathMNIST and OrganAMNIST are below the diagonal line and those of the other datasets approximately align with the diagonal line. Notably, CONFINE is able to achieve validity by using the softmax layer as the feature extraction layer for PathMNIST and OrganAMNIST, but these prior methods cannot. The prior methods also lack class-conditional coverage: only some class have coverage curves above the diagonal line.

\begin{figure}
    \centering
    \includegraphics[width=1\linewidth]{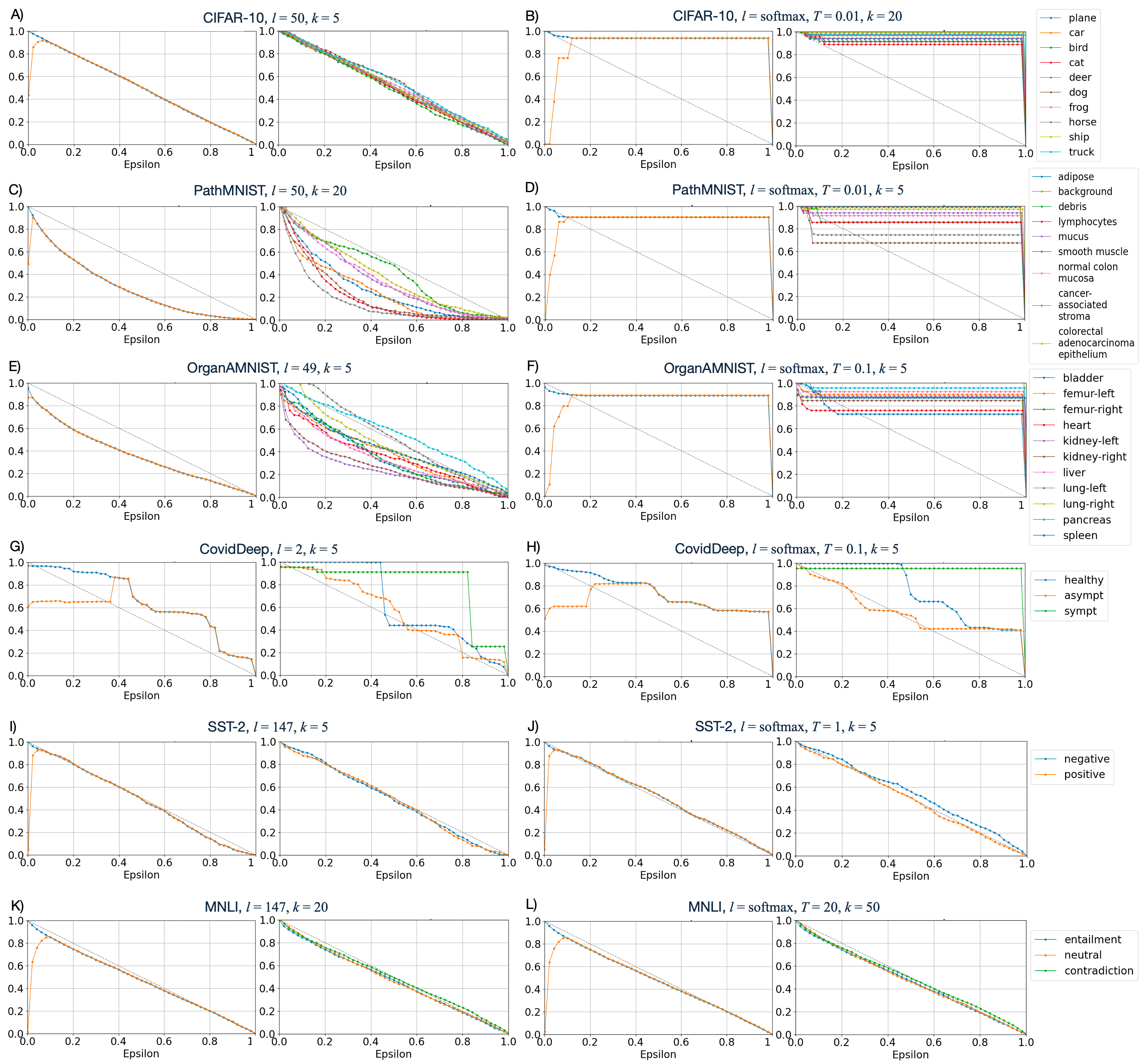}
    \caption{Overall coverage (left) and class-wise coverage (right) curves with the class-wise split of the calibration set for all six datasets using different hyperparameters.}
    \label{fig:coverage-split-all}
\end{figure}

\subsection{Coverage Curves of CONFINE With Class-wise Split}\label{app:coverage-split-all}

Fig.~\ref{fig:coverage-split-all} illustrates the overall coverage and correct efficiency curves and class-wise coverage curves for all six datasets with the class-wise split of the calibration set. Similarly, other layers before the classification head with different hyperparameters have results that look similar to Fig.~\ref{fig:coverage-split-all} A, C, E, G, I, K, and the softmax layer with other hyperparameters have results that look similar to Fig.~\ref{fig:coverage-split-all} B, D, F, H, J, L.

With the class-wise split, for CIFAR-10, CovidDeep, SST-2, and MNLI, all class coverage curves become higher and align with the diagonal line, demonstrating CONFINE's ability to achieve class-conditional coverage (Fig.~\ref{fig:coverage-split-all} A, B, G, H, I, J, K, L). For PathMNIST and OrganAMNIST, however, the class-wise coverage curves of layers before softmax are still below the diagonal line due to non-exchangeability -- the overall coverage curves, which are from class-wise coverage curves aggregated together, sit below the diagonal line (Fig.~\ref{fig:coverage-split-all} C, E). For these two datasets, we can use the softmax layer to achieve validity and class-conditional coverage (Fig.~\ref{fig:coverage-split-all} D, F).

\subsection{Effect of Class-wise Split}\label{app:res-classwise}
Class-wise split generally causes CONFINE's overall accuracy and top correct efficiency to slightly decrease except for CovidDeep where they remain the same (Table~\ref{tab:accuracy-classwise}). The performance losses are very subtle: overall accuracy drops by 1.65\% for OrganAMNIST and drops by less than 0.4\% for the other datasets; overall top correct efficiency drops by about 3\% and 5\% for PathMNIST and OrganAMNIST, respectively, drops by less than 0.6\% for MNLI and CIFAR-10, and even increases by about 0.6\% for SST-2. This is because under the class-wise split of the calibration set, the data size used for calculating the $p$-values becomes much smaller, so data noisiness and bias would have larger effects on the $p$-value calculation. The class-averaged accuracy also does not always increase because many of the datasets have well-balanced classes, meaning their class-averaged accuracy is very close to the overall accuracy.

\begin{table*}[tbp]
\caption{Class-wise split causes slight decreases in overall accuracy and top correct efficiency. CONFINE: without the calibration set class-wise split; CONFINE-classwise: with the calibration set class-wise split. -A: hyperparameter search for highest test accuracy; -C: hyperparameter search for highest top correct efficiency. Hyperparameter details are given in Appendix~\ref{app:hparams}.}
\label{tab:accuracy-classwise}
\vskip 0.15in
\begin{center}
\small
\begin{tabularx}{\linewidth}{@{\extracolsep{\fill}} llcccc}
\toprule
Dataset & Method & Test Acc & \begin{tabular}{@{}c@{}}Class-averaged\\Test Acc\end{tabular} & Top Corr Effi \\
\midrule
\multirow{5}{*}{CIFAR-10} & Original NN & 93.92 & 93.92 & -- \\
 & CONFINE-A & \textbf{94.37} ($\uparrow$) & \textbf{94.37} & 93.72 \\
 & CONFINE-C & 93.92 (--) & 93.92 & \textbf{93.92} \\
 & CONFINE-classwise-A & 94.08 ($\uparrow$) & 94.08 & 91.21 \\
 & CONFINE-classwise-C & 93.89 ($\downarrow$) & 93.89 & 93.73 \\
\midrule
\multirow{5}{*}{PathMNIST} & Original NN & 90.65 & 87.55 & -- \\
 & CONFINE-A & \textbf{94.22} ($\uparrow$) & 91.84 & 93.44 \\
 & CONFINE-C & 94.09 ($\uparrow$) & 91.71 & \textbf{93.91}\\
 & CONFINE-classwise-A & 94.14 ($\uparrow$) & \textbf{92.20} & 90.15 \\
 & CONFINE-classwise-C & 94.01 ($\uparrow$) & 91.93 & 90.97 \\
\midrule
\multirow{5}{*}{OrganAMNIST} & Original NN & 91.78 & 90.43 & -- \\
 & CONFINE-A & \textbf{94.78} ($\uparrow$) & \textbf{94.30} & 92.94\\
 & CONFINE-C & 94.45 ($\uparrow$) & 93.84 & \textbf{94.04} \\
 & CONFINE-classwise-A & 93.13 ($\uparrow$) & 93.24 & 87.01 \\
 & CONFINE-classwise-C & 92.64 ($\uparrow$) & 92.55 & 89.00 \\
 \midrule
\multirow{3}{*}{CovidDeep} & Original NN & \textbf{98.07} & \textbf{98.23} & -- \\
  & CONFINE-A/-C & \textbf{98.07} (--) & \textbf{98.23} & \textbf{98.07} \\
 & CONFINE-classwise-A/-C & \textbf{98.07} (--) & \textbf{98.23} & \textbf{98.07} \\
\midrule
\multirow{5}{*}{SST-2} & Original NN & 93.23 & 93.24 & -- \\
 & CONFINE-A & \textbf{93.69} ($\uparrow$) & \textbf{93.70} & 92.43\\
 & CONFINE-C & 93.57 ($\uparrow$) & 93.59 & 92.66 \\
 & CONFINE-classwise-A & 93.35 ($\uparrow$) & 93.37 & 92.43 \\
 & CONFINE-classwise-C & 93.23 (--) & 93.24 & \textbf{93.23} \\
\midrule
\multirow{5}{*}{MNLI} & Original NN & 86.56 & 86.45 & -- \\
 & CONFINE-A & \textbf{86.67} ($\uparrow$) & \textbf{86.56} & 85.71\\
 & CONFINE-C & 86.60 ($\uparrow$) & 86.50 & \textbf{85.79} \\
 & CONFINE-classwise-A & 86.35 ($\downarrow$) & 86.37 & 85.10 \\
 & CONFINE-classwise-C & 86.30 ($\downarrow$) & 86.32 & 85.20 \\
\bottomrule
\end{tabularx}
\end{center}
\vskip -0.1in
\end{table*}

\section{Computational Complexity Analysis}\label{sec:comp-complexity}
An important limitation of CONFINE is its trade-off of computational efficiency for interpretability, just like a lot of other conformal prediction methods, due to the fact that the entire proper training set is needed when calculating the nonconformity score of a new test sample.

In the pre-processing step, every instance in the proper training set and the calibration set needs to go through one neural network inference for feature extraction, on which nearest neighbors computation is performed. If $d$ is the size of the extracted feature vector, pre-processing takes $O((N_t+N_c)T_f + N_cdN_t)$ time, where $N_t$ is the size of the proper training set, $N_c$ is the size of the calibration set, and $T_f$ is the time taken for one inference of neural network $f$. This pre-processing step is carried out only once and the feature vectors of the proper training set are stored, which takes $O(dN_t)$ space. Then, running CONFINE for a single test sample requires one neural network inference and one nearest neighbors computation, taking $O(T_f + dN_t)$ time. Compared with the original neural network inference, which only takes $O(T_f)$ time, $O(dN_t)$ extra space and time are required by CONFINE to provide interpretability.

The computational complexity is affected by the size of the proper training set, $N_t$, and the dimension of the extracted feature, $d$. The latter depends on which layer of the neural network we use. For example, the softmax layer would have $d=C$, where $C$ is the number of classes, the layer before the classifier head of ResNet18 has $d=512$ and that of RoBERTa-base has $d=768$.

\end{appendices}

\end{document}